\PassOptionsToPackage{table}{xcolor}
\documentclass[10pt,logo,copyright]{nvidiatechreport}
\usepackage[authoryear,round]{natbib}

\usepackage[utf8]{inputenc}
\usepackage[T1]{fontenc}

\usepackage{parskip}
\usepackage{url}
\usepackage{xurl}
\usepackage{booktabs}
\usepackage{amsfonts}
\usepackage{nicefrac}
\usepackage{microtype}
\usepackage{xcolor}
\usepackage[dvipsnames]{xcolor}
\usepackage{graphicx}
\usepackage{animate}
\usepackage{subcaption}
\usepackage{tabularx}
\usepackage{makecell}
\usepackage{adjustbox}
\usepackage{setspace}
\newcolumntype{M}[1]{>{\centering\arraybackslash}m{#1}}
\usepackage{float}
\usepackage{tikz}
\usetikzlibrary{positioning,shapes,arrows}
\usepackage{amsmath,amsfonts,bm, bbm,leftindex}
\usepackage{multirow}
\usepackage{comment}
\usepackage{gensymb}
\usepackage{lipsum}
\usetikzlibrary{arrows.meta, positioning, fit}
\usepackage[para]{threeparttable}
\usepackage{tikz}
\usetikzlibrary{tikzmark}

\usepackage[most]{tcolorbox}
\usepackage{fancyvrb}
\usepackage{fvextra}
\usepackage{dashrule}

\DeclareRobustCommand{\apicode}[1]{\begingroup\Urlmuskip=0mu plus 1mu\urlstyle{tt}\nolinkurl{#1}\endgroup}
\usepackage{calc}
\usepackage{wrapfig}
\usepackage{listings}
\usepackage{etoc}
\usepackage[nameinlink]{cleveref}

\crefname{equation}{Eq.}{Eqs.}
\crefname{figure}{Fig.}{Figs.}
\crefname{section}{Sec.}{Sec.}
\crefname{appendix}{App.}{App.}
\crefname{table}{Tab.}{Tabs.}

\definecolor{darkred}{rgb}{0.7, 0.0, 0.0}

\newcommand{\ours}{SpatialClaw}

\lstdefinestyle{pythonStyle}{%
  language=Python,
  basicstyle=\ttfamily\scriptsize,
  keywordstyle=\color{blue!70!black},
  commentstyle=\color{gray!70!black}\itshape,
  stringstyle=\color{red!60!black},
  showstringspaces=false,
  breaklines=true,
  columns=fullflexible,
  upquote=true,
  aboveskip=4pt,
  belowskip=2pt,
}

\definecolor{deltapos}{RGB}{0,130,0}
\definecolor{deltaneg}{RGB}{200,30,30}
\definecolor{ngreen}{HTML}{76B900}
\colorlet{bestcell}{ngreen!60}
\colorlet{secondcell}{ngreen!20}
\newcommand{\up}[1]{\,{\scriptsize\color{deltapos}(+#1)}}
\newcommand{\dn}[1]{\,{\scriptsize\color{deltaneg}($-$#1)}}
\newcommand{\eq}[1]{\,{\scriptsize\color{gray}(+#1)}}
\newcommand{\paragrapht}{\paragraph}

\newtcolorbox{findingbox}[1][]{%
  enhanced,
  colback=gray!8,
  colframe=gray!60,
  fonttitle=\bfseries,
  coltitle=black,
  colbacktitle=gray!30,
  sharp corners,
  boxrule=0.6pt,
  top=6pt,
  bottom=6pt,
  left=6pt,
  right=6pt,
  title={#1}
}

\newcounter{findingcounter}
\newcommand{\finding}[2]{%
    \refstepcounter{findingcounter}%
    \begin{tcolorbox}[
        colback=white!90!gray,
        colframe=ngreen,
        arc=2pt,
        boxsep=3pt,
        left=10pt,
        right=10pt,
        top=0pt,
        bottom=0pt,
        boxrule=0.8pt,
        enhanced jigsaw
    ]
        \textbf{Finding~\arabic{findingcounter}.}~\textbf{#1} #2
    \end{tcolorbox}
    \vspace{-0.1cm}
}

\title{\ours{}:\\Rethinking Action Interface for Agentic Spatial Reasoning}
\newcommand{\headertitle}{\ours{}: Rethinking Action Interface for Agentic Spatial Reasoning}

\author{Seokju Cho$^{1}$, Ryo Hachiuma, Abhishek Badki, Hang Su, Byung-Kwan Lee, Chan Hee Song, Sifei Liu,
Subhashree Radhakrishnan, Seungryong Kim$^{1}$, Yu-Chiang Frank Wang and Min-Hung Chen\\
{\Affilfont \textbf{Links}:~\href{https://github.com/NVlabs/SpatialClaw}{Code}~\textperiodcentered~\href{https://spatialclaw.github.io/}{Project Page}}\\
{\Affilfont NVIDIA}}
\authornotes{\textsuperscript{1}\,Affiliated with KAIST. Work done during Seokju Cho's internship at NVIDIA.}

\hypersetup{
  pdftitle={SpatialClaw: Rethinking Action Interface for Agentic Spatial Reasoning},
  pdfauthor={Seokju Cho, Ryo Hachiuma, Abhishek Badki, Hang Su, Byung-Kwan Lee, Chan Hee Song, Sifei Liu, Subhashree Radhakrishnan, Seungryong Kim, Yu-Chiang Frank Wang, Min-Hung Chen}
}

\begin{document}

\etocdepthtag.toc{mainpaper}

\maketitle
\vspace{-6.5mm}

%

\definecolor{catA}{HTML}{5A9000} 
\definecolor{catB}{HTML}{3068B4} 
\definecolor{catC}{HTML}{8C3C8C} 
\definecolor{catD}{HTML}{B56B1C} 
\definecolor{catE}{HTML}{28828C} 
\definecolor{oursLine}{HTML}{5A9000}
\definecolor{oursFill}{HTML}{76B900}
\definecolor{stLine}{HTML}{3264B4}
\definecolor{pyLine}{HTML}{E8821E}
\definecolor{ntLine}{HTML}{7A7A7A}

\newcommand{\spatialradar}[2]{%
\begin{tikzpicture}[font=\sffamily]
  \def\Rmax{3.2}\def\sc{0.032}%
  \pgfmathsetmacro{\step}{360/#1}%
  \pgfmathtruncatemacro{\Nmm}{#1-1}%
  \foreach \v in {20,40,60,80,100}{\pgfmathsetmacro{\rr}{\v*\sc}\draw[gray!28,line width=0.3pt] (0,0) circle (\rr);}%
  \foreach \nm/\cat/\anc/\nt/\st/\py/\lr [count=\i from 0] in {#2}{%
    \pgfmathsetmacro{\ang}{90-\step*\i}%
    \pgfmathsetmacro{\rnt}{\nt*\sc}\pgfmathsetmacro{\rst}{\st*\sc}\pgfmathsetmacro{\rpy}{\py*\sc}%
    \draw[gray!22,line width=0.3pt] (0,0) -- (\ang:\Rmax);%
    \coordinate (O\i) at (\ang:2.56);\coordinate (N\i) at (\ang:\rnt);%
    \coordinate (S\i) at (\ang:\rst);\coordinate (P\i) at (\ang:\rpy);%
    \node[anchor=\anc,text=\cat,inner sep=1.3pt,font=\scriptsize] at (\ang:\lr) {\nm};%
  }%
  \fill[oursFill,fill opacity=0.18] (O0) \foreach \i in {1,...,\Nmm}{ -- (O\i)} -- cycle;%
  \draw[ntLine,line width=0.8pt,dash pattern=on 3pt off 2pt,opacity=0.9] (N0) \foreach \i in {1,...,\Nmm}{ -- (N\i)} -- cycle;%
  \draw[pyLine,line width=1.0pt] (P0) \foreach \i in {1,...,\Nmm}{ -- (P\i)} -- cycle;%
  \draw[stLine,line width=1.0pt] (S0) \foreach \i in {1,...,\Nmm}{ -- (S\i)} -- cycle;%
  \draw[oursLine,line width=1.7pt] (O0) \foreach \i in {1,...,\Nmm}{ -- (O\i)} -- cycle;%
  \foreach \i in {0,...,\Nmm}{\fill[ntLine] (N\i) circle (1.0pt);\fill[pyLine] (P\i) circle (1.1pt);\fill[stLine] (S\i) circle (1.1pt);}%
  \foreach \i in {0,...,\Nmm}{\fill[oursFill] (O\i) circle (1.5pt);\draw[oursLine,line width=0.5pt] (O\i) circle (1.5pt);}%
\end{tikzpicture}}

\begin{figure}[H]
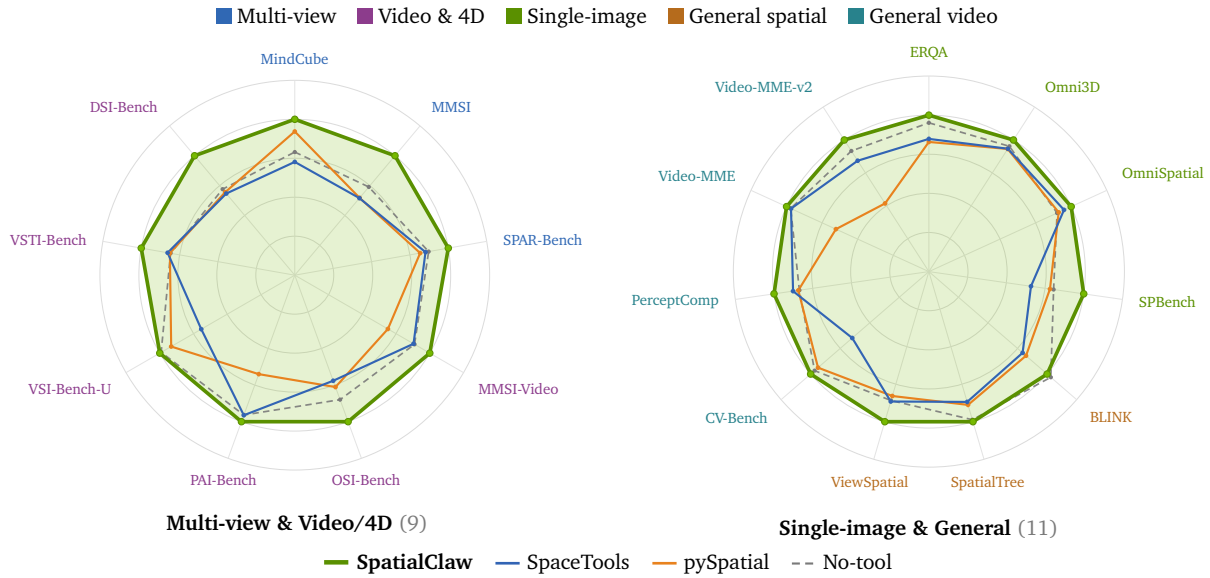

\centering
{\footnotesize
\tikz\fill[catB] (0,0) rectangle (0.2,0.2);~Multi-view\quad
\tikz\fill[catC] (0,0) rectangle (0.2,0.2);~Video \& 4D\quad
\tikz\fill[catA] (0,0) rectangle (0.2,0.2);~Single-image\quad
\tikz\fill[catD] (0,0) rectangle (0.2,0.2);~General spatial\quad
\tikz\fill[catE] (0,0) rectangle (0.2,0.2);~General video}\par
\vspace{5pt}
\begin{minipage}[c]{0.46\linewidth}
\centering
\resizebox{\linewidth}{!}{\spatialradar{9}{%
  MindCube/catB/{south}/63.19/58.13/73.74/3.42,
  MMSI/catB/{south west}/59.10/51.62/51.77/3.42,
  SPAR-Bench/catB/{west}/69.76/68.12/65.34/3.42,
  MMSI-Video/catC/{north west}/70.96/70.38/55.19/3.42,
  OSI-Bench/catC/{north}/67.97/57.66/61.10/3.42,
  PAI-Bench/catC/{north}/76.36/76.48/54.04/3.42,
  VSI-Bench-U/catC/{north east}/79.18/55.42/73.24/3.42,
  VSTI-Bench/catC/{east}/64.73/66.27/65.09/3.42,
  DSI-Bench/catC/{south east}/57.62/54.69/55.58/3.42%
}}\\[2pt]
{\footnotesize\textbf{Multi-view \& Video/4D} \textcolor{ntLine}{(9)}}
\end{minipage}%
\hspace{0.035\linewidth}%
\begin{minipage}[c]{0.46\linewidth}
\centering
\resizebox{\linewidth}{!}{\spatialradar{11}{%
  ERQA/catA/{south}/76.09/67.86/66.30/3.42,
  Omni3D/catA/{south west}/76.17/74.70/74.55/3.42,
  OmniSpatial/catA/{south west}/72.08/75.97/72.96/3.42,
  SPBench/catA/{west}/64.44/52.75/62.57/3.42,
  BLINK/catD/{north west}/82.51/63.43/65.72/3.42,
  SpatialTree/catD/{north}/78.95/69.46/71.04/3.42,
  ViewSpatial/catD/{north}/68.70/69.24/66.31/3.42,
  CV-Bench/catE/{north east}/77.34/51.86/75.01/3.42,
  PerceptComp/catE/{east}/66.91/70.18/67.45/3.42,
  Video-MME/catE/{south east}/77.71/77.51/52.26/3.42,
  Video-MME-v2/catE/{south east}/73.33/67.39/41.44/3.42%
}}\\[2pt]
{\footnotesize\textbf{Single-image \& General} \textcolor{ntLine}{(11)}}
\end{minipage}

\vspace{4pt}
{\footnotesize
\raisebox{1.5pt}{\tikz{\draw[oursLine,line width=1.7pt] (0,0) -- (0.32,0);}}~\textbf{\ours{}}\quad
\raisebox{1.5pt}{\tikz{\draw[stLine,line width=1.0pt] (0,0) -- (0.32,0);}}~SpaceTools\quad
\raisebox{1.5pt}{\tikz{\draw[pyLine,line width=1.0pt] (0,0) -- (0.32,0);}}~pySpatial\quad
\raisebox{1.5pt}{\tikz{\draw[ntLine,line width=0.8pt,dash pattern=on 3pt off 2pt] (0,0) -- (0.32,0);}}~No-tool}
\caption{\textbf{\ours{} improves spatial reasoning across the board.}
Per-benchmark accuracy on \textbf{20 spatial reasoning benchmarks} (Gemma\,4-31B
backbone), split into two panels by task category. Each axis is individually
rescaled so \ours{} traces the constant-radius ring. Baselines are
SpaceTools-Toolshed~\citep{chen2025spacetools}, pySpatial~\citep{luo2026pyspatial}, and a no-tool backbone.}
\label{fig:radar}
\end{figure}
\vspace{-8.5mm}

\begin{abstract}
    Spatial reasoning, the ability to determine where objects are, how they relate, and how they move in 3D, remains a fundamental challenge for vision-language models (VLMs).
    Tool-augmented agents attempt to address this by augmenting VLMs with specialist perception modules, yet their effectiveness is bounded by the \emph{action interface} through which those tools are invoked.
    In this work, we study how the design of this interface shapes the agent's capacity for open-ended spatial reasoning.
    Existing spatial agents either employ single-pass code execution, which commits to a full analysis strategy before any intermediate result is observed, or rely on a structured tool-call interface that often offers less flexibility for freely composing operations or tailoring the analysis to each task. Both designs offer limited flexibility for open-ended, complex 3D/4D spatial reasoning.
    We therefore propose \textbf{\ours{}}, a training-free framework for spatial reasoning that adopts code as the action interface.
    \ours{} maintains a stateful Python kernel pre-loaded with input frames and a suite of perception and geometry primitives, letting a VLM-backed agent write one executable cell per step conditioned on all prior outputs, enabling the agent to flexibly compose and manipulate perception results and adapt its analysis to both intermediate text and visual observations and the demands of each problem.
    Evaluated across \textbf{20 spatial reasoning benchmarks} spanning a broad range of static and dynamic 3D/4D spatial reasoning tasks, \ours{} achieves 59.9\% average accuracy, outperforming the recent spatial agent by \textbf{+11.2 points}, with consistent gains across six VLM backbones from two model families \emph{without any benchmark- or model-specific adaptation}.
\end{abstract}

\abscontent
\section{Introduction}
\label{sec:intro}

\begin{figure}[t]
    \centering
    \includegraphics[width=\linewidth]{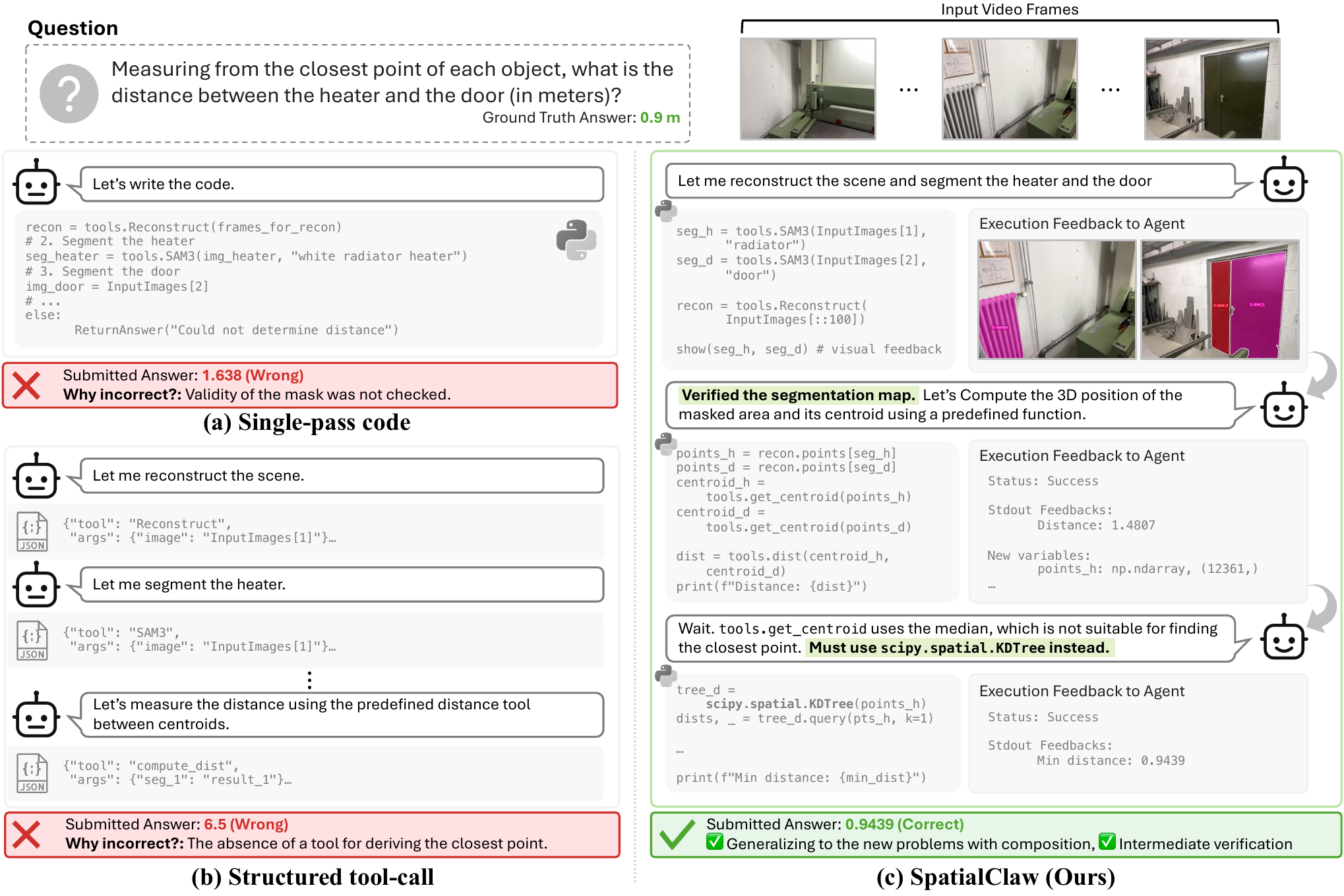}
    \caption{\textbf{\ours{} studies code as the action interface for spatial reasoning.} Three action interfaces on the same question. (a) A single-pass program chooses a complete computation before seeing its intermediate outputs. (b) A structured tool interface exposes common operations through structured commands (\emph{e.g.}, JSON, XML). (c) \ours{} writes Python in a persistent kernel, renders intermediate evidence, and revises the measurement before answering.}
    \label{fig:teaser}
    \vspace{-4em}
\end{figure}

Spatial reasoning, the ability to determine where objects are, how they move, and how they relate in three dimensions, is foundational to visual perception.
Questions such as \emph{``Is the car moving toward the camera?''}, \emph{``Which object is closest to the table?''}, and \emph{``Did the person turn left or right?''} are effortless for humans, yet remain challenging for vision-language models (VLMs), including state-of-the-art models~\citep{yang2024think,chen2025eagle2.5,qwen3.5,gemma4}.
Answering them requires composing multiple types of evidence, including depth, camera pose, and temporal correspondence, into a coherent geometric argument, and current VLMs struggle to perform this structured analysis reliably from pixels alone~\citep{chen2024spatialvlm,cheng2024spatialrgpt,cho20264dpqa}.

A natural response is to augment VLMs with specialist perception tools~\citep{Gupta2022VisProg,surismenon2023vipergpt,shen2023hugginggpt,lu2026octotools,chen2025spacetools,han2025tiger}, such as detectors~\citep{carion2020end}, segmenters~\citep{ravi2024sam,carion2025sam}, and depth or pose estimators~\citep{wang2025pi}, that produce perceptual outputs that VLMs would otherwise have to approximate from visual appearance alone.
The capability of a tool-augmented agent, however, depends not only on which tools are available, but also on the \emph{action interface} through which those tools are invoked.
The action interface specifies how tools are invoked and their outputs represented, which intermediate states are observable between steps, and whether the agent can revise its reasoning in light of those observations before committing to an answer.

Prior tool-augmented spatial agents have primarily relied on one of two action interfaces.
In \emph{single-pass code execution}, the agent writes a complete Python program and runs it once~\citep{luo2026pyspatial,marsili2025visual} (Fig.~\ref{fig:teaser}a), enabling flexible tool invocation but requiring a complete analysis strategy to be committed before any intermediate result has been observed.
In \emph{structured tool-calls}, the agent selects from a list of named tools, fills typed arguments, and consumes typed return values~\citep{chen2025geometrically,chen2025spacetools,ropero2026riemind}, offering limited support for composing and synthesizing tool outputs through external scientific libraries such as \texttt{numpy} or \texttt{scipy} to perform task-specific computations that emerge only at test time (Fig.~\ref{fig:teaser}b).

In practice, neither design readily supports the open-ended, compositional reasoning that complex 3D/4D spatial reasoning tasks often demand.
We therefore argue that the appropriate action interface for spatial reasoning should treat code not as a one-shot program or a dispatch interface for pre-registered tools, but as an \emph{orchestration space} in which the agent sequences, composes, and diagnoses the perception tools it has access to. In light of this, we propose \textbf{\ours{}}, a training-free framework that instantiates this principle as code as the action interface (Fig.~\ref{fig:teaser}c).

In \ours{}, perception tools are Python callables and their outputs, including masks, depth maps, camera geometry, and trajectories, are ordinary Python variables; the kernel preserves this state independently across turns, so any object produced at one step remains available for composition, inspection, and revision at all subsequent steps.
While structured tool-calls and natural language reasoning can partially approximate this behavior, code generation provides a more expressive interface for adapting perception to the task, especially in spatial reasoning, where the required computations often cannot be anticipated by a fixed API.
A new spatial analysis is not a new API entry, but a new composition of perception tools (e.g., segmentation~\citep{carion2025sam} or reconstruction~\citep{depthanything3}) and numerical primitives (e.g., \texttt{numpy}~\citep{harris2020array}, \texttt{scipy}~\citep{2020SciPy-NMeth}) assembled across steps in response to intermediate evidence.
\ours{} pre-loads a persistent Python kernel with input frames, perception tools, and scientific libraries, and coordinates a VLM-backed agent through an iterative loop of planning, code execution, and feedback assembly, guided by a unified system prompt that encodes general principles of spatial reasoning rather than task-specific instructions.

We evaluate \ours{} on 20 spatial reasoning benchmarks spanning metric distance, camera and object motion, multi-view geometry, temporal reasoning, and spatial planning, to name a few.
\ours{} achieves 59.9\% average accuracy across all 20 benchmarks, outperforming the recent spatial agent~\citep{chen2025spacetools} by \textbf{+11.2 points} (Fig.~\ref{fig:radar}).
Gains are largest on dynamic 4D video reasoning and multi-view inference, precisely the categories that demand chained geometric computation across frames and viewpoints, where no pre-specified tool call captures the required composition.
Strikingly, these gains generalize out of the box: \ours{} delivers consistent improvements across the large majority of benchmarks and backbone models tested, spanning two distinct model families, Qwen~\citep{qwen3.5,qwen3.6-27b,qwen36_35b_a3b} and Gemma4~\citep{gemma4}, with parameters ranging from 27B to 397B, \emph{without any modification} to the system prompt, tool set, or benchmark-specific engineering, confirming that the expressive action interface itself drives the gains rather than model-specific tuning.
We further present a controlled comparison of three action interfaces and ablation studies, demonstrating that our interface achieves the strongest generalization through flexible composition of perception primitives and that the gains persist even when all pre-defined utility wrappers are removed.

Our contributions are as follows:
\begin{itemize}[leftmargin=1.5em,itemsep=1pt,topsep=1pt]
  \item \textbf{Rethinking the action interface for spatial reasoning agents.}
    We argue that the design of the action interface is a critical factor in agent capability in spatial reasoning, and introduce \ours{}, a training-free agent that instantiates this principle by replacing single-pass code execution and structured tool-call interfaces with a persistent, multi-turn Python kernel, turning each step into an opportunity to compose perception outputs with numerical primitives, inspect intermediate state, and revise the analysis.
  \item \textbf{Comprehensive evaluation that validates action interface design.}
    We establish an extensive evaluation spanning single-image spatial reasoning, multi-view spatial reasoning, general spatial reasoning, video spatial \& 4D reasoning, and general video understanding, on which \ours{} achieves consistent improvement over baselines across most benchmarks and transfers across backbone models without any benchmark- or model-specific adaptation, providing a comprehensive empirical comparison of spatial agents.
\end{itemize}
\section{Action Interfaces for Spatial Reasoning Agents}
\label{sec:interfaces}
\label{sec:code}

We characterize tool-augmented spatial agents by their \emph{action interface}, that is, the medium through which they acquire, inspect, and transform visual evidence.
When no interface is used, the VLM performs spatial reasoning by generating a chain-of-thought rationale in natural language directly over the input images~\citep{yang2024think,chen2025eagle2.5,qwen3.5,gemma4}, without invoking any external computation or acquiring intermediate evidence.
Beyond this tool-free case, prior spatial agents have primarily relied on one of two action interfaces: single-pass code execution, or predefined API calls through structured tool-calls.

\paragraph{Single-pass code.}
The agent writes a complete Python program, runs it once, and uses it to call perception modules and process their outputs~\citep{luo2026pyspatial,marsili2025visual}.
The program can express task-specific computations, but it must commit to a complete strategy before observing any intermediate mask, depth map, plot, or runtime error.
Retry mechanisms may repair a syntax or runtime failure, but intermediate visual evidence does not participate in the reasoning loop.

\paragraph{Structured tool-calls.}
The agent invokes perception modules by name, passing typed inputs and receiving typed outputs through predefined structured commands~\citep{chen2025geometrically,chen2025spacetools,ropero2026riemind}, e.g., JSON or XML.
While this exposes a clean interface for invoking perception, compositions that can only be determined at test time, such as chaining a depth estimate with a segmentation mask in a way specific to the question, are difficult to express within the predefined command schema.

\paragraph{Motivation.} These considerations motivate a third interface centered on code as a medium for flexible iterative composition based on intermediate evidence.
Building on the rapid advances in code generation capabilities of modern LLMs, \ours{} lets the agent write one Python cell per step and execute it in a persistent kernel, conditioning each subsequent step on the resulting text, variables, errors, and visualizations; that is, we interpret code as the action interface.
Code generation provides a more general mechanism for adapting to the task than either approach: computations such as finding the nearest object via \texttt{scipy.spatial.KDTree} or estimating a surface plane via RANSAC~\citep{fischler1981random} emerge naturally from the problem at hand, without needing to be anticipated by a predefined interface.
Spatial results such as depth maps, camera geometry, and temporal trajectories persist as ordinary Python variables across steps, remaining available for inspection, composition, and correction throughout execution.
Fig.~\ref{fig:teaser} provides a comparison of these three interfaces.

\section{SpatialClaw}
\label{sec:method}

\ours{} realizes this code as the action interface through a persistent Python workspace for spatial reasoning.
Rather than exposing a larger menu of spatial tools, it lets the agent turn visual evidence into inspectable spatial analyses that can be composed and revised over multiple steps.
For each example, \ours{} creates a persistent Python kernel pre-loaded with the input frames and a set of primitives spanning perception modules (e.g., depth estimation, segmentation) and scientific libraries (e.g., NumPy~\citep{harris2020array}, SciPy~\citep{2020SciPy-NMeth}, Matplotlib~\citep{Hunter:2007}), and asks a VLM-backed agent to write one executable code cell at a time.
Each cell can create masks, reconstructions, plots, or numeric summaries, and the resulting program state remains available to later cells.
The next action is conditioned on text output, variable summaries, and rendered intermediate images from previous steps.
The remainder of this section formalizes these two components: the persistent kernel workspace that maintains shared program state across execution steps (\S\ref{sec:workspace}), and the outer agentic loop that coordinates planning, code generation, feedback assembly, and answer submission (\S\ref{sec:loop}).

\begin{figure}[t]
    \centering
    \includegraphics[width=\linewidth]{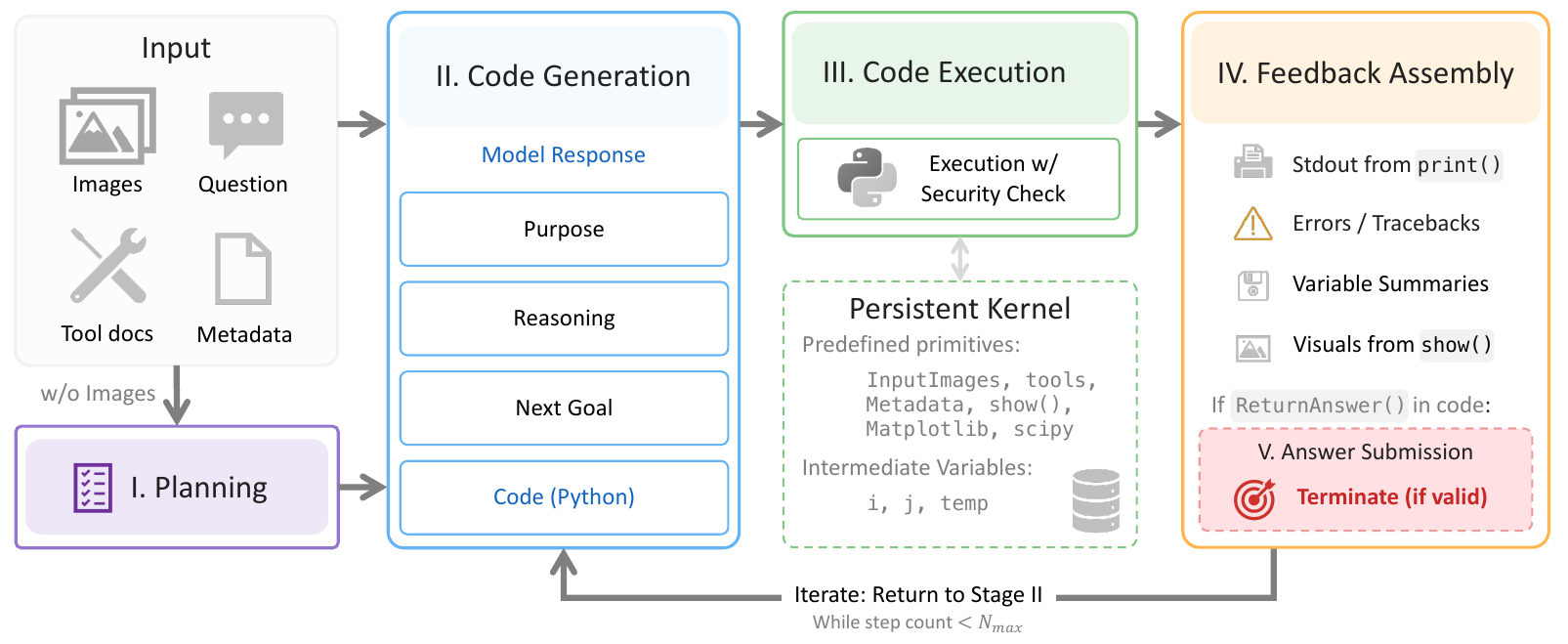}
    \caption{\textbf{Agentic loop for iterative code execution.} 
    \ours{} wraps a persistent kernel in a five-stage loop. A planner receives the question and tool documentation but not the images, and produces an analysis plan. The main agent generates a Python cell executed in the persistent kernel. Feedback comprising stdout, variable summaries, and images registered via \texttt{show()} is appended to the model context. The loop continues until the agent submits an answer with \texttt{ReturnAnswer()} or the step count has reached the predefined maximum $N_{\text{max}}$.}
    \label{fig:method_loop}
\end{figure}

\subsection{Persistent Kernel Workspace}
\label{sec:workspace}

The workspace is initialized once for an example and discarded after the system terminates.
It keeps all intermediate results as ordinary Python variables, so any object produced during execution, including segmentation masks, depth maps, numeric arrays, and rendered plots, remains accessible to subsequent code cells.
As a result, the agent does not need to choose the complete analysis in advance.
It can start with a coarse computation, inspect the result, and refine the analysis against the observed evidence. The kernel exposes six public entry points.
\begin{enumerate}[leftmargin=2em, itemsep=1pt, topsep=1pt]
    \item \texttt{InputImages} contains the sampled frames or images.
    \item \texttt{Metadata} contains frame rate, duration, and frame indices for video inputs, enabling the agent to reason about temporal structure when questions reference specific timestamps.
    \item \texttt{tools} exposes evidence-producing perception and geometry primitives.
    \texttt{Reconstruct} wraps Depth Anything 3~\citep{depthanything3} and returns per-frame depth, camera intrinsics, camera extrinsics, and dense point maps.
    \texttt{SAM3}~\citep{carion2025sam} produces image or video masks from text, point, or box prompts.
    The \texttt{tools} module also includes lightweight utilities for mask operations, geometric computations, and visualizations.
    These utility functions simplify common operations such as mask manipulation and geometric computation, but the agent is not limited to what they provide; any operation can also be implemented directly over arrays or through the scientific libraries (\texttt{NumPy}, \texttt{SciPy}, \texttt{Matplotlib}) available in the execution environment. Full descriptions of all tools and utilities are provided in Appendix~\ref{app:tools}.
    \item \texttt{show(\ldots)} registers an image to be embedded directly into the agent's context at the next step, allowing the agent to visually inspect intermediate results such as masks, depth maps, or annotated frames before deciding what to do next. Figures rendered via \texttt{matplotlib} (\texttt{plt.show()}) are captured in the same manner and included in the agent's next observation.
    \item \texttt{vlm} dispatches queries to a separate VLM session, returning the response as plain text without embedding images into the main agent's context.
    \texttt{vlm.locate(\ldots)} performs visual grounding, returning bounding boxes.
    \texttt{vlm.ask\_with\_thinking(\ldots)} handles questions that fall outside the perception toolset, such as those requiring commonsense reasoning or artistic style recognition. 
    Further details on the prompts used for each VLM session are given in Appendix~\ref{app:prompts}.
    \item \texttt{ReturnAnswer(\ldots)} submits a candidate answer.
\end{enumerate}
Together, these entry points form a self-contained computational environment: the agent reads observations through \texttt{InputImages} and \texttt{Metadata}, builds up spatial evidence via \texttt{tools}, inspects intermediate results through \texttt{show(\ldots)}, queries a separate VLM session when needed via \texttt{vlm}, and commits a final answer through \texttt{ReturnAnswer(\ldots)}.
How the agent sequences these actions across multiple steps is governed by the outer loop described next.

\subsection{Spatial Reasoning Loop: Code, Inspect, and Revise}
\label{sec:loop}

The persistent kernel provides a runtime environment for the action interface, but does not by itself prescribe when to plan, when to act on the observed evidence, and when to commit to an answer.
We therefore wrap the kernel in a five-stage outer loop (Fig.~\ref{fig:method_loop}) covering planning, code generation, code execution, feedback assembly, and answer submission.
The loop follows standard agent control structures and serves as a stable scaffold under which the construct, inspect, and revise pattern operates consistently across examples.
The loop prescribes the order and termination conditions of each stage, but not how the agent should reason spatially within them; that reasoning discipline is encoded in the system prompt. Details of the agent system are provided in \S\ref{app:system}.

\paragrapht{Disciplined reasoning via system prompt.}
\ours{} imposes structure on the open-ended Python action space through a unified system prompt, without any benchmark-specific engineering.
The prompt defines both the runtime objects available in the kernel and the verification discipline the agent is required to follow.
The prompt encourages the agent to treat spatial conclusions as claims that should be cross-checked against multiple evidence sources where possible.
For instance, the agent is guided to resolve the relevant frame of reference, prefer metric computation over pixel-level impressions for geometric questions, and visually inspect tool outputs such as masks and annotated objects.
It is also encouraged to sanity-check numerical magnitudes and consider any apparent disagreements between visualizations and computed quantities before invoking \texttt{ReturnAnswer()}.
Because the prompt encodes general principles rather than few-shot examples or task-specific templates, the same configuration is applied across all benchmarks and backbone models without modification; more prompt details are given in \S\ref{app:prompts}.
With this discipline in place, the loop proceeds through five stages as follows:

\paragrapht{Stage I: Planning.}
The planner runs in a separate LLM session, isolated from the main agent's execution context, with its own dedicated system prompt, and is invoked once per sample prior to execution.
Unlike the main agent's prompt, which focuses on general spatial reasoning discipline, the planner's prompt forbids executable code and pre-conclusion phrases, and instead instructs the planner to outline the analysis steps and the evidence to be acquired.
It receives the question, metadata, and tool documentation, but not the input frames for efficiency.
The resulting plan is appended to the main agent's system prompt before execution begins, providing the agent with a structured plan to follow.

\paragrapht{Stage II: Code generation.}
The main VLM-backed agent receives the question, the plan, the execution trajectory, and any images shown in previous steps.
At each step, it produces a structured response comprising \textit{purpose}, \textit{reasoning}, \textit{next goal}, and \textit{code} fields, formatted in markdown.
The \textit{code} field is a Python cell that implements the next analysis action, which can call perception primitives, transform their outputs, render intermediate evidence, or submit an answer via \texttt{ReturnAnswer()}.

\paragrapht{Stage III: Code execution.}
Before execution, a static checker parses the code's abstract syntax tree (AST) to reject disallowed modules and unsafe builtins without running the code; on failure, the framework injects a concise traceback into the context and prompts the agent to revise the code.
The validated cell is then executed in the persistent kernel.

\paragrapht{Stage IV: Feedback assembly.}
The next model context receives the standard output produced by the executed cell (e.g., values printed via \texttt{print()}), concise tracebacks for any failed execution, summaries of newly created variables including their type, length, and size, and any images registered through \texttt{show()} for visual inspection.
Because the kernel persists, later code can reuse earlier masks, reconstructions, plots, and partial analyses without recomputing them.
The agent can therefore revise the analysis after seeing a wrong mask, an implausible trajectory, or a depth distribution that does not support the intended comparison. 
This feedback is appended to the model context, making intermediate results accessible at any subsequent step.
The loop then returns to Stage II unless the agent has submitted an answer or the step count has reached the predefined maximum $N_{\text{max}}$.

\paragrapht{Stage V: Answer submission.}
When the agent determines that it has collected sufficient evidence, it submits an answer with \texttt{ReturnAnswer()}.
The loop terminates once the submitted answer conforms to the expected format for the question type (e.g., multiple choice, numerical, or free-form text). Otherwise, the loop continues to the next step. 

\begin{table*}[t]
\caption{\textbf{Main results across 20 spatial reasoning benchmarks.} For each backbone, we compare a no-tool baseline (thinking mode without tool access) with our \ours{} agent framework. 
}
\label{tab:main_results}
\centering
\setlength{\tabcolsep}{3.5pt}
\renewcommand{\arraystretch}{1.15}

{\small
\resizebox{\textwidth}{!}{%
\begin{tabular}{ll cccc ccc ccc}
\toprule
\textbf{Model} & \textbf{Method}
  & \multicolumn{4}{c}{\textit{Single-image spatial reasoning}}
  & \multicolumn{3}{c}{\textit{Multi-view spatial reasoning}}
  & \multicolumn{3}{c}{\textit{General spatial reasoning}} \\
\cmidrule(lr){3-6} \cmidrule(lr){7-9} \cmidrule(lr){10-12}
 &
  & \rotatebox{80}{ERQA~\citeyearpar{team2025gemini}}
  & \rotatebox{80}{Omni3D~\citeyearpar{marsili2025visual}}
  & \rotatebox{80}{OmniSpatial~\citeyearpar{jia2025omnispatial}}
  & \rotatebox{80}{SPBench~\citeyearpar{xu2025spbench}}
  & \rotatebox{80}{MindCube~\citeyearpar{yin2025mindcube}}
  & \rotatebox{80}{MMSI~\citeyearpar{yang2025mmsi}}
  & \rotatebox{80}{SPAR-Bench~\citeyearpar{zhang2025from}}
  & \rotatebox{80}{BLINK~\citeyearpar{fu2024blink}}
  & \rotatebox{80}{SpatialTree~\citeyearpar{xiao2025spatialtree}}
  & \rotatebox{80}{ViewSpatial~\citeyearpar{li2025viewspatial}} \\
\midrule
\multirow{2}{*}{Qwen3.5-397B-A17B~\citep{qwen3.5}}
  & No-tool & 63.0 & 57.0 & 62.1 & 54.0 & 58.3 & 46.2 & 57.2 & 72.9 & 65.1 & 57.3 \\
  & \ours{}  & 62.0\dn{1.0} & 55.1\dn{1.9} & 62.8\up{0.7} & 63.3\up{9.3} & 66.2\up{7.9} & 49.7\up{3.5} & 66.1\up{8.9} & 76.7\up{3.8} & 60.0\dn{5.1} & 57.3\eq{0.0} \\
\midrule
\multirow{2}{*}{Qwen3.5-122B-A10B~\citep{qwen3.5}}
  & No-tool & 61.0 & 55.1 & 43.3 & 52.2 & 49.7 & 42.7 & 52.6 & 73.2 & 61.6 & 52.8 \\
  & \ours{}  & 57.5\dn{3.5} & 52.3\dn{2.8} & 61.9\up{18.6} & 63.6\up{11.4} & 61.0\up{11.3} & 45.3\up{2.6} & 57.6\up{5.0} & 74.6\up{1.4} & 57.8\dn{3.8} & 54.5\up{1.7} \\
\midrule
\multirow{2}{*}{Qwen3.6-35B-A3B~\citep{qwen36_35b_a3b}}
  & No-tool & 57.2 & 51.2 & 57.0 & 52.4 & 45.4 & 38.0 & 47.8 & 70.6 & 58.3 & 52.0 \\
  & \ours{}  & 58.2\up{1.0} & 54.0\up{2.8} & 59.9\up{2.9} & 63.0\up{10.6} & 61.9\up{16.5} & 45.9\up{7.9} & 58.4\up{10.6} & 74.1\up{3.5} & 60.3\up{2.0} & 52.9\up{0.9} \\
\midrule
\multirow{2}{*}{Qwen3.6-27B~\citep{qwen3.6-27b}}
& No-tool & 59.5 & 55.9 & 60.8 & 58.0 & 51.2 & 40.9 & 55.0 & 70.5 & 62.6 & 53.9 \\
& \ours{}  & 61.5\up{2.0} & 57.7\up{1.8} & 62.8\up{2.0} & 64.8\up{6.8} & 70.1\up{18.9} & 53.9\up{13.0} & 68.1\up{13.1} & 77.4\up{6.9} & 64.6\up{2.0} & 59.1\up{5.2} \\
\midrule
\multirow{2}{*}{Gemma4-31B~\citep{gemma4}}
  & No-tool & 58.3 & 51.7 & 57.3 & 55.1 & 57.5 & 37.9 & 55.2 & 75.7 & 59.9 & 51.7 \\
  & \ours{}  & 61.3\up{3.0} & 54.3\up{2.6} & 63.6\up{6.3} & 68.4\up{13.3} & 72.8\up{15.3} & 51.3\up{13.4} & 63.3\up{8.1} & 73.4\dn{2.3} & 60.7\up{0.8} & 60.2\up{8.5} \\
\midrule
\multirow{2}{*}{Gemma4-26B-A4B~\citep{gemma4}}
  & No-tool & 56.0 & 44.8 & 56.9 & 45.1 & 47.8 & 31.5 & 48.3 & 70.2 & 53.7 & 53.1 \\
  & \ours{}  & 56.0\eq{0.0} & 48.9\up{4.1} & 57.1\up{0.2} & 61.9\up{16.8} & 64.0\up{16.2} & 42.7\up{11.2} & 63.1\up{14.8} & 71.5\up{1.3} & 57.9\up{4.2} & 58.6\up{5.5} \\
\bottomrule
\end{tabular}%
}}

{\small
\resizebox{\textwidth}{!}{%
\begin{tabular}{ll cccccc cccc >{\columncolor{gray!15}}c}
\toprule
\textbf{Model} & \textbf{Method}
  & \multicolumn{6}{c}{\textit{Video spatial \& 4D reasoning}}
  & \multicolumn{4}{c}{\textit{General video understanding}}
  & \multicolumn{1}{c}{\textit{Average}} \\
\cmidrule(lr){3-8} \cmidrule(lr){9-12} \cmidrule(lr){13-13}
 &
  & \rotatebox{80}{MMSI-Video~\citeyearpar{lin2025mmsivideo}}
  & \rotatebox{80}{OSI-Bench~\citeyearpar{wu2025osibench}}
  & \rotatebox{80}{PAI-Bench~\citeyearpar{zhou2025paibench}}
  & \rotatebox{80}{VSI-Bench-U~\citeyearpar{brown2025shortcuts}}
  & \rotatebox{80}{VSTI-Bench~\citeyearpar{fan2025vlm}}
  & \rotatebox{80}{DSI-Bench~\citeyearpar{zhang2025dsibench}}
  & \rotatebox{80}{CV-Bench~\citeyearpar{zhu2025cvbench}}
  & \rotatebox{80}{PerceptComp~\citeyearpar{li2026perceptioncomp}}
  & \rotatebox{80}{Video-MME~\citeyearpar{fu2025video}}
  & \rotatebox{80}{Video-MME-v2~\citeyearpar{videommev2_2026}}
  & \rotatebox{80}{\textbf{Avg.}} \\
\midrule
\multirow{2}{*}{Qwen3.5-397B-A17B~\citep{qwen3.5}}
  & No-tool & 40.8 & 37.3 & 73.3 & 58.8 & 60.4 & 49.4 & 71.7 & 42.1 & 77.5 & 41.1 & \textbf{57.3} \\
  & \ours{}  & 43.6\up{2.8} & 45.6\up{8.3} & 71.6\dn{1.7} & 56.6\dn{2.2} & 67.5\up{7.1} & 65.8\up{16.4} & 72.9\up{1.2} & 43.0\up{0.9} & 77.0\dn{0.5} & 45.7\up{4.6} & \textbf{60.4}\up{3.1} \\
\midrule
\multirow{2}{*}{Qwen3.5-122B-A10B~\citep{qwen3.5}}
  & No-tool & 37.9 & 37.1 & 69.8 & 55.7 & 54.7 & 47.5 & 72.1 & 41.6 & 75.8 & 38.2 & \textbf{53.7} \\
  & \ours{}  & 38.3\up{0.4} & 41.4\up{4.3} & 65.6\dn{4.2} & 50.6\dn{5.1} & 63.3\up{8.6} & 66.4\up{18.9} & 72.2\up{0.1} & 41.1\dn{0.5} & 72.6\dn{3.2} & 40.2\up{2.0} & \textbf{56.9}\up{3.2} \\
\midrule
\multirow{2}{*}{Qwen3.6-35B-A3B~\citep{qwen36_35b_a3b}}
  & No-tool & 35.6 & 35.5 & 67.6 & 54.2 & 58.8 & 47.2 & 71.1 & 40.4 & 74.3 & 37.0 & \textbf{52.6} \\
  & \ours{}  & 38.5\up{2.9} & 41.9\up{6.4} & 63.8\dn{3.8} & 51.7\dn{2.5} & 64.2\up{5.4} & 64.1\up{16.9} & 74.0\up{2.9} & 42.0\up{1.6} & 75.0\up{0.7} & 40.1\up{3.1} & \textbf{57.2}\up{4.6} \\
\midrule
\multirow{2}{*}{Qwen3.6-27B~\citep{qwen3.6-27b}}
& No-tool & 37.1 & 36.8 & 69.8 & 55.4 & 60.5 & 48.0 & 70.6 & 39.4 & 76.7 & 36.7 & \textbf{55.0} \\
& \ours{}  & 47.0\up{9.9} & 47.9\up{11.1} & 72.1\up{2.3} & 57.9\up{2.5} & 70.3\up{9.8} & 68.1\up{20.1} & 75.1\up{4.5} & 46.5\up{7.1} & 79.4\up{2.7} & 49.7\up{13.0} & \textbf{62.7}\up{7.7} \\
\midrule
\multirow{2}{*}{Gemma4-31B~\citep{gemma4}}
  & No-tool & 36.9 & 35.6 & 65.0 & 48.0 & 54.7 & 45.3 & 69.8 & 36.8 & 74.8 & 40.7 & \textbf{53.4} \\
  & \ours{}  & 41.6\up{4.7} & 41.9\up{6.3} & 68.1\up{3.1} & 48.5\up{0.5} & 67.6\up{12.9} & 62.9\up{17.6} & 72.2\up{2.4} & 44.0\up{7.2} & 77.0\up{2.2} & 44.4\up{3.7} & \textbf{59.9}\up{6.5} \\
\midrule
\multirow{2}{*}{Gemma4-26B-A4B~\citep{gemma4}}
  & No-tool & 32.5 & 31.1 & 59.1 & 30.1 & 53.1 & 41.1 & 66.3 & 34.8 & 69.5 & 35.6 & \textbf{48.0} \\
  & \ours{}  & 32.8\up{0.3} & 40.0\up{8.9} & 59.0\dn{0.1} & 38.5\up{8.4} & 61.7\up{8.6} & 61.1\up{20.0} & 67.8\up{1.5} & 36.3\up{1.5} & 69.6\up{0.1} & 37.3\up{1.7} & \textbf{54.3}\up{6.3} \\
\bottomrule
\end{tabular}%
}}
\end{table*}

\begin{table}[t]
\centering

\newsavebox{\actionLBox}\newsavebox{\actionRBox}
\newlength{\actionLcap}\newlength{\actionRcap}

\sbox{\actionLBox}{%
  \setlength{\tabcolsep}{2pt}\renewcommand{\arraystretch}{1.05}\scriptsize
  \begin{tabular}[t]{l cccc}
  \toprule
  \rowcolor{white}
    & No-tool   & Single-Pass & Structured & \ours{}      \\
  \rowcolor{white}\multirow{-2}{*}{\textbf{Benchmark}}
    & Baseline  & Code        & Tool-Call  & (Ours)       \\
  \midrule
  \multicolumn{5}{l}{\textit{Single-image spatial reasoning}} \\
  \quad ERQA~\citep{team2025gemini}              & 58.3 & 58.3 & \cellcolor{secondcell}\underline{59.0} & \cellcolor{bestcell}\textbf{61.3} \\
  \quad Omni3D~\citep{marsili2025visual}        & 51.7 & 48.2 & \cellcolor{bestcell}\textbf{55.7} & \cellcolor{secondcell}\underline{54.3} \\
  \quad OmniSpatial~\citep{jia2025omnispatial}  & 57.3 & \cellcolor{secondcell}\underline{60.0} & 59.6 & \cellcolor{bestcell}\textbf{63.6} \\
  \quad SPBench~\citep{xu2025spbench}           & 55.1 & 58.2 & \cellcolor{secondcell}\underline{60.5} & \cellcolor{bestcell}\textbf{68.4} \\
  \midrule
  \multicolumn{5}{l}{\textit{Multi-view spatial reasoning}} \\
  \quad MindCube~\citep{yin2025mindcube}        & 57.5 & 57.5 & \cellcolor{secondcell}\underline{62.4} & \cellcolor{bestcell}\textbf{72.8} \\
  \quad MMSI~\citep{yang2025mmsi}               & 37.9 & 42.3 & \cellcolor{secondcell}\underline{43.0} & \cellcolor{bestcell}\textbf{51.3} \\
  \quad SPAR-Bench~\citep{zhang2025from}        & 55.2 & \cellcolor{secondcell}\underline{61.1} & 58.7 & \cellcolor{bestcell}\textbf{63.3} \\
  \midrule
  \multicolumn{5}{l}{\textit{Video spatial \& 4D reasoning}} \\
  \quad MMSI-Video~\citep{lin2025mmsivideo}     & 36.9 & \cellcolor{secondcell}\underline{37.1} & 35.1 & \cellcolor{bestcell}\textbf{41.6} \\
  \quad OSI-Bench~\citep{wu2025osibench}        & 35.6 & 36.8 & \cellcolor{secondcell}\underline{40.3} & \cellcolor{bestcell}\textbf{41.9} \\
  \quad PAI-Bench~\citep{zhou2025paibench}      & 65.0 & 64.2 & \cellcolor{secondcell}\underline{65.6} & \cellcolor{bestcell}\textbf{68.1} \\
  \quad VSI-Bench-U~\citep{brown2025shortcuts}  & 48.0 & 47.8 & \cellcolor{bestcell}\textbf{50.1} & \cellcolor{secondcell}\underline{48.5} \\
  \quad VSTI-Bench~\citep{fan2025vlm}           & 54.7 & \cellcolor{secondcell}\underline{64.2} & 63.5 & \cellcolor{bestcell}\textbf{67.6} \\
  \quad DSI-Bench~\citep{zhang2025dsibench}     & 45.3 & 57.9 & \cellcolor{secondcell}\underline{58.4} & \cellcolor{bestcell}\textbf{62.9} \\
  \midrule
  \multicolumn{5}{l}{\textit{General spatial reasoning}} \\
  \quad BLINK~\citep{fu2024blink}               & \cellcolor{bestcell}\textbf{75.7} & \cellcolor{secondcell}\underline{73.9} & \cellcolor{secondcell}\underline{73.9} & 73.4 \\
  \quad SpatialTree~\citep{xiao2025spatialtree} & \cellcolor{secondcell}\underline{59.9} & 58.9 & 57.7 & \cellcolor{bestcell}\textbf{60.7} \\
  \quad ViewSpatial~\citep{li2025viewspatial}   & 51.7 & 52.0 & \cellcolor{secondcell}\underline{55.5} & \cellcolor{bestcell}\textbf{60.2} \\
  \midrule
  \multicolumn{5}{l}{\textit{General video understanding}} \\
  \quad CV-Bench~\citep{zhu2025cvbench}          & 69.8 & 71.3 & \cellcolor{bestcell}\textbf{73.6} & \cellcolor{secondcell}\underline{72.2} \\
  \quad PerceptComp~\citep{li2026perceptioncomp}& 36.8 & 39.2 & \cellcolor{secondcell}\underline{42.5} & \cellcolor{bestcell}\textbf{44.0} \\
  \quad Video-MME~\citep{fu2025video}           & 74.8 & 74.6 & \cellcolor{secondcell}\underline{75.8} & \cellcolor{bestcell}\textbf{77.0} \\
  \quad Video-MME-v2~\citep{videommev2_2026}    & 40.7 & 40.2 & \cellcolor{secondcell}\underline{44.0} & \cellcolor{bestcell}\textbf{44.4} \\
  \midrule
  \textbf{Average} (20 bench.) & 53.4 & 55.2 & \cellcolor{secondcell}\underline{56.7} & \cellcolor{bestcell}\textbf{59.9} \\
  \bottomrule
  \end{tabular}%
}
\sbox{\actionRBox}{%
  \setlength{\tabcolsep}{2pt}\renewcommand{\arraystretch}{1.05}\scriptsize
  \begin{tabular}[t]{cccc}
  \toprule
  \rowcolor{white}VADAR                       & pySpatial                & SpaceTools Toolshed                          & \ours{} \\
  \rowcolor{white}\citep{marsili2025visual}   & \citep{luo2026pyspatial} & \citep{chen2025spacetools} & (Ours)  \\
  \midrule
  \multicolumn{4}{c}{\strut} \\
  33.3 & 50.8 & \cellcolor{secondcell}\underline{52.0} & \cellcolor{bestcell}\textbf{61.3} \\
  44.6 & 50.6 & \cellcolor{secondcell}\underline{50.7} & \cellcolor{bestcell}\textbf{54.3} \\
  42.4 & 58.0 & \cellcolor{secondcell}\underline{60.4} & \cellcolor{bestcell}\textbf{63.6} \\
  41.6 & \cellcolor{secondcell}\underline{53.5} & 45.1 & \cellcolor{bestcell}\textbf{68.4} \\
  \midrule
  \multicolumn{4}{c}{\strut} \\
  --$^{*}$ & \cellcolor{secondcell}\underline{67.1} & 52.9 & \cellcolor{bestcell}\textbf{72.8} \\
  --$^{*}$ & \cellcolor{secondcell}\underline{33.2} & 33.1 & \cellcolor{bestcell}\textbf{51.3} \\
  --$^{*}$ & 51.7 & \cellcolor{secondcell}\underline{53.9} & \cellcolor{bestcell}\textbf{63.3} \\
  \midrule
  \multicolumn{4}{c}{\strut} \\
  --$^{*}$ & 28.7 & \cellcolor{secondcell}\underline{36.6} & \cellcolor{bestcell}\textbf{41.6} \\
  --$^{*}$ & \cellcolor{secondcell}\underline{32.0} & 30.2 & \cellcolor{bestcell}\textbf{41.9} \\
  --$^{*}$ & 46.0 & \cellcolor{secondcell}\underline{65.1} & \cellcolor{bestcell}\textbf{68.1} \\
  --$^{*}$ & \cellcolor{secondcell}\underline{44.4} & 33.6 & \cellcolor{bestcell}\textbf{48.5} \\
  --$^{*}$ & 55.0 & \cellcolor{secondcell}\underline{56.0} & \cellcolor{bestcell}\textbf{67.6} \\
  --$^{*}$ & \cellcolor{secondcell}\underline{43.7} & 43.0 & \cellcolor{bestcell}\textbf{62.9} \\
  \midrule
  \multicolumn{4}{c}{\strut} \\
  --$^{*}$ & \cellcolor{secondcell}\underline{60.3} & 58.2 & \cellcolor{bestcell}\textbf{73.4} \\
  --$^{*}$ & \cellcolor{secondcell}\underline{53.9} & 52.7 & \cellcolor{bestcell}\textbf{60.7} \\
  --$^{*}$ & 49.9 & \cellcolor{secondcell}\underline{52.1} & \cellcolor{bestcell}\textbf{60.2} \\
  \midrule
  \multicolumn{4}{c}{\strut} \\
  --$^{*}$ & \cellcolor{secondcell}\underline{67.7} & 46.8 & \cellcolor{bestcell}\textbf{72.2} \\
  --$^{*}$ & 37.1 & \cellcolor{secondcell}\underline{38.6} & \cellcolor{bestcell}\textbf{44.0} \\
  --$^{*}$ & 50.3 & \cellcolor{secondcell}\underline{74.6} & \cellcolor{bestcell}\textbf{77.0} \\
  --$^{*}$ & 23.0 & \cellcolor{secondcell}\underline{37.4} & \cellcolor{bestcell}\textbf{44.4} \\
  \midrule
  --$^{*}$ & 47.8 & \cellcolor{secondcell}\underline{48.7} & \cellcolor{bestcell}\textbf{59.9} \\
  \bottomrule
  \end{tabular}%
}

\setlength{\actionLcap}{(\linewidth-8pt) * \ratio{\wd\actionLBox}{\wd\actionLBox+\wd\actionRBox}}
\setlength{\actionRcap}{(\linewidth-8pt) * \ratio{\wd\actionRBox}{\wd\actionLBox+\wd\actionRBox}}

\begin{minipage}[t]{\actionLcap}
\caption{\textbf{Action interface comparison.} All variants use the same toolset. ``Structured Tool-Call'' exposes the tools through a JSON command interface~\citep{chen2025spacetools}. ``Single-Pass Code'' generates one complete program before execution.}
\label{tab:action_interface}
\end{minipage}%
\hspace{8pt}%
\begin{minipage}[t]{\actionRcap}
\caption{\textbf{Comparison with other spatial agents.} All
use the same Gemma4-31B~\citep{gemma4} backbone model. 
 $^{*}$:~video or multi-image not supported.}
\label{tab:comparison_agents}
\end{minipage}

\resizebox{\linewidth}{!}{\usebox{\actionLBox}\hspace{8pt}\usebox{\actionRBox}}

\end{table}

\begin{figure*}[t]
    \centering
    \includegraphics[width=\linewidth]{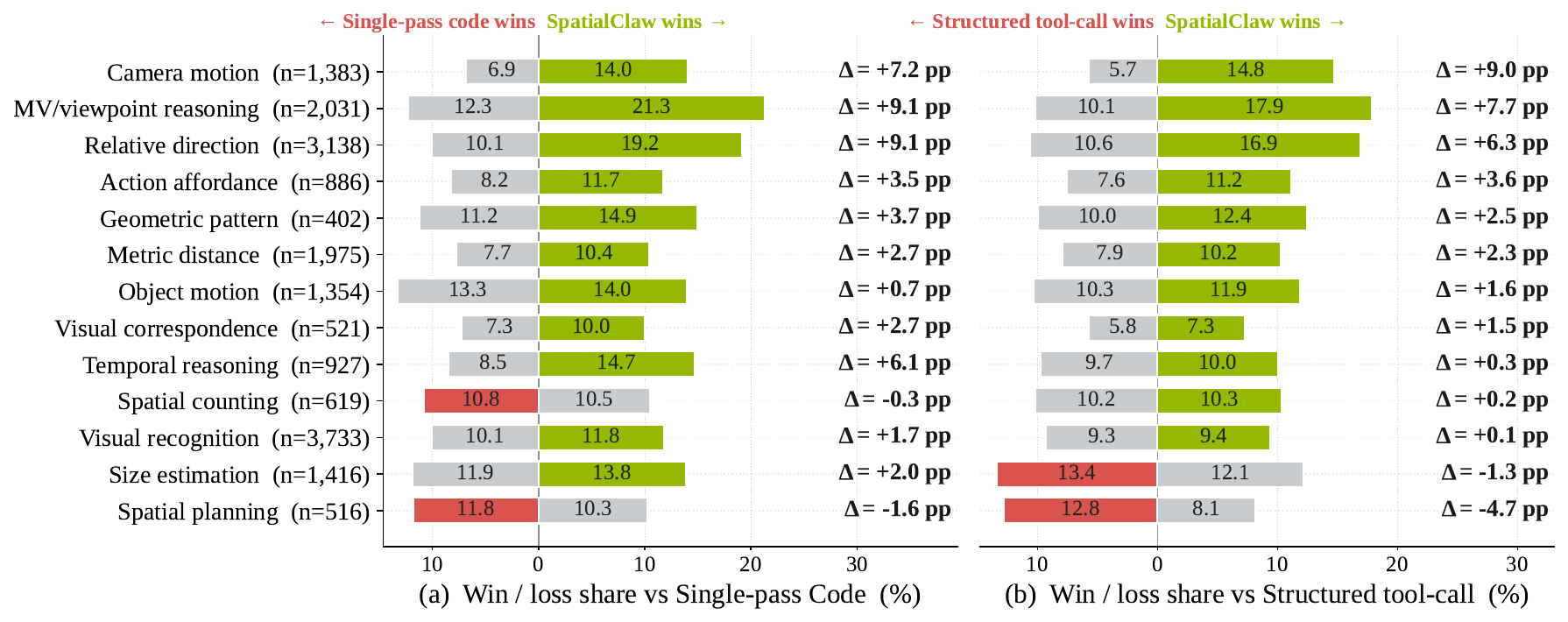}
    \caption{\textbf{Pairwise win/loss margin of \ours{} over baselines across 13 meta-categories.} 
    \ours{} outperforms both (a) Structured tool-call and (b) Single-pass Code in 11/13 categories. 
    The largest gains concentrate in categories that demand multi-step geometric composition.}
    \label{fig:pairwise_winloss_per_meta}
\end{figure*}

\section{Results}

\paragrapht{Evaluation setup.}
We evaluate on 20 spatial reasoning benchmarks spanning single-image spatial reasoning~\citep{team2025gemini,marsili2025visual,jia2025omnispatial,xu2025spbench}, multi-view spatial reasoning~\citep{yin2025mindcube,yang2025mmsi,zhang2025from}, general spatial reasoning~\citep{fu2024blink,xiao2025spatialtree,li2025viewspatial}, video spatial \& 4D reasoning~\citep{lin2025mmsivideo,wu2025osibench,zhou2025paibench,brown2025shortcuts,fan2025vlm,zhang2025dsibench}, and general video understanding~\citep{zhu2025cvbench,li2026perceptioncomp,fu2025video,videommev2_2026}. 
Evaluation details are provided in \S\ref{app:benchmarks}.
We evaluate across several open-source VLM backbones, including the 122B-A10B and 397B-A17B variants of Qwen3.5~\citep{qwen3.5}, the 35B-A3B and 27B variants of Qwen3.6~\citep{qwen3.6-27b}, and the 31B and 26B-A4B variants of Gemma4~\citep{gemma4}. 
Due to the large number of benchmarks and backbones, we cap evaluation at $1,000$ samples for benchmarks exceeding that size; otherwise, all samples are used. $N_{max}=30$ is used for all benchmarks.
For all results in Tab.~\ref{tab:main_results} and Tab.~\ref{tab:action_interface}, the same hyperparameters, system prompts, and perception tools are used across all benchmarks and backbone models; further details are in the supplementary material.

\paragrapht{Benchmark results.}
\label{sec:main_results}
Tab.~\ref{tab:main_results} reports results across 20 spatial reasoning benchmarks, spanning a broad range of tasks including video reasoning.
\ours{} consistently improves over the no-tool baseline across all six backbone models, with the most pronounced gains on video spatial \& 4D reasoning (e.g., DSI-Bench, avg.\ $+18.3\%p$) and multi-view spatial reasoning (e.g., MindCube, avg.\ $+14.3\%p$), categories that benefit most from iterative multi-step geometric computation across frames or viewpoints.
Notably, these gains hold consistently across models ranging from 26B to 397B parameters without any modification, suggesting that our design generalizes to future models \textit{without model-specific tuning}.

\paragrapht{Action interface comparison.}
\label{sec:abl_action}

Tab.~\ref{tab:action_interface} compares \ours{} against two alternative action interfaces that share the same toolset and system prompt, differing only in the response format and the system prompt section describing the action interface. 
We also include a \emph{No-tool} baseline that receives visual inputs and reasons directly without tool access. Implementation details for each baseline are provided in \S\ref{app:baselines}.
\emph{Single-Pass Code} moves to open code, but asks the model to generate one complete program before seeing execution feedback;
\emph{Structured Tool-Call} adds external perception through a JSON command interface. The results show that \ours{} consistently outperforms the other two action interfaces across all benchmarks, with the largest gains on tasks that require multi-step geometric composition. 
These results validate the generalizability of \ours{}'s action interface for spatial reasoning.

\begin{table*}[t]
  \caption{\textbf{Analysis on tools of \ours{}.}
  We ablate two design choices: No utility function, keeping only the perception tools (SAM3/DA3) and removing perception tools and keeping only the utility functions. We additionally report a no-tool baseline (backbone with thinking only). Each variant is evaluated on subsampled examples (500 per benchmark) across 15 benchmarks due to computation constraints. Gemma4-26B-A4B is used as the backbone for its small size.
  }
  \label{tab:ablation}
  \centering
  \setlength{\tabcolsep}{2.5pt}
  \renewcommand{\arraystretch}{1.15}
  \small
  \newcolumntype{C}{>{\centering\arraybackslash}p{1.6em}}
  \newlength{\genvidcolwidth}\setlength{\genvidcolwidth}{5.5em}    
  \newlength{\avgcolwidth}\setlength{\avgcolwidth}{3em}        
  \newcolumntype{Y}{>{\centering\arraybackslash}p{\genvidcolwidth}}
  \newcolumntype{Z}{>{\centering\arraybackslash}p{\avgcolwidth}}
  \resizebox{\textwidth}{!}{%
  \begin{tabular}{l CCCC !{\hskip 8pt} CCC !{\hskip 8pt} CCC !{\hskip 8pt} CCCC !{\hskip 8pt} Y !{\hskip 8pt} >{\columncolor{gray!15}}Z}
  \toprule
  \textbf{Variant}
    & \multicolumn{4}{c}{\itshape\shortstack[c]{Single-image\\spatial reasoning}}
    & \multicolumn{3}{c}{\itshape\shortstack[c]{Multi-view\\spatial reasoning}}
    & \multicolumn{3}{c}{\itshape\shortstack[c]{General\\spatial reasoning}}
    & \multicolumn{4}{c}{\itshape\shortstack[c]{Video spatial\\\& 4D reasoning}}
    & \multicolumn{1}{c}{\itshape\shortstack[c]{General video\\understanding}}
    & \multicolumn{1}{c}{\itshape\shortstack[c]{\\Average}} \\
  \cmidrule(lr){2-5} \cmidrule(lr){6-8} \cmidrule(lr){9-11} \cmidrule(lr){12-15} \cmidrule(lr){16-16} \cmidrule(lr){17-17}
   & \rotatebox{90}{ERQA~\citeyearpar{team2025gemini}}
   & \rotatebox{90}{Omni3D~\citeyearpar{marsili2025visual}}
   & \rotatebox{90}{OmniSpatial~\citeyearpar{jia2025omnispatial}}
   & \rotatebox{90}{SPBench~\citeyearpar{xu2025spbench}}
   & \rotatebox{90}{MindCube~\citeyearpar{yin2025mindcube}}
   & \rotatebox{90}{MMSI~\citeyearpar{yang2025mmsi}}
   & \rotatebox{90}{SPAR-Bench~\citeyearpar{zhang2025from}}
   & \rotatebox{90}{BLINK~\citeyearpar{fu2024blink}}
   & \rotatebox{90}{SpatialTree~\citeyearpar{xiao2025spatialtree}}
   & \rotatebox{90}{ViewSpatial~\citeyearpar{li2025viewspatial}}
   & \rotatebox{90}{OSI-Bench~\citeyearpar{wu2025osibench}}
   & \rotatebox{90}{PAI-Bench~\citeyearpar{zhou2025paibench}}
   & \rotatebox{90}{VSTI-Bench~\citeyearpar{fan2025vlm}}
   & \rotatebox{90}{DSI-Bench~\citeyearpar{zhang2025dsibench}}
   & \rotatebox{90}{CV-Bench~\citeyearpar{zhu2025cvbench}}
   & \rotatebox{90}{\textbf{Avg.}} \\
  \midrule
  \ours{} (Full)
    & \underline{56.0} & \underline{36.6} & 57.2 & \textbf{65.2} & \underline{67.2} & \textbf{45.1} & \textbf{67.7} & \textbf{71.9} & \textbf{37.8} & \textbf{60.6} & \textbf{44.2} & \textbf{56.2} & \underline{58.3} & \underline{61.0} & \textbf{68.9} & \textbf{56.9} \\
  \midrule
  \textbf{(I)} No utility functions (e.g., \texttt{tools.Mask}, \texttt{tools.Geometry})
    & 52.5 & \textbf{37.0} & \underline{61.6} & \underline{62.4} & \textbf{69.0} & \underline{43.1} & \underline{66.1} & \textbf{71.9} & 35.1 & \underline{60.4} & \underline{42.8} & 53.0 & \textbf{59.9} & \textbf{62.0} & \underline{68.7} & \underline{56.4} \\
  \textbf{(II)} No perception tools (no SAM3/DA3)
    & \textbf{58.0} & 33.8 & \textbf{62.2} & 58.4 & 49.0 & 35.5 & 53.8 & \underline{70.1} & 33.1 & 56.8 & 34.0 & 55.2 & 53.6 & 48.8 & \textbf{68.9} & 51.4 \\
  \midrule
  No-tool baseline
    & 55.8 & 32.8 & 56.6 & 47.0 & 47.6 & 31.5 & 50.8 & 69.1 & \underline{35.5} & 54.8 & 35.0 & \underline{56.0} & 49.6 & 43.6 & 65.4 & 48.7 \\
  \bottomrule
  \end{tabular}}
\end{table*}

\paragrapht{Quantitative comparison with other spatial agent methods.}
\label{sec:abl_comparison}
Tab.~\ref{tab:comparison_agents} compares \ours{} against recent spatial agent methods~\citep{marsili2025visual,luo2026pyspatial,chen2025spacetools} using the same Gemma4-31B~\citep{gemma4} backbone for all methods, using the official implementation of each method. 
Here, VADAR and pySpatial fall into the category of single-pass code, while SpaceTools uses the structured tool-call interface.
VADAR does not support video or multi-image inputs, so the corresponding entries are left blank.

Overall, \ours{} outperforms all baselines across all benchmarks, achieving the largest margin over SpaceTools ($+11.2\%p$ on average). Among the baselines, SpaceTools ranks highest, consistent with Tab.~\ref{tab:action_interface}, which shows that structured tool-calls generally outperform single-pass code. \ours{} further improves over both action interfaces across all evaluated setups.

We also note that the baselines do not consistently improve over the no-tool baseline. For SpaceTools, we hypothesize that the method is designed to be fine-tuned via reinforcement learning, which may limit its zero-shot generalization. Other baselines may similarly have been optimized for narrower task categories, which could partly explain their limited performance across the diverse benchmarks evaluated here.

\section{Analysis and Insights}
\label{sec:analysis}
In this section, we investigate \emph{why} \ours{}'s action interface improves spatial reasoning beyond what structured alternatives can achieve.

\finding{\ours{} generalizes across diverse spatial reasoning tasks even without pre-defined utility tools.}

Tab.~\ref{tab:ablation} reports an ablation study examining how the composition of the tool set affects performance. In variant \textbf{(I)}, we remove all utility wrappers (e.g., \texttt{tools.Mask}, \texttt{tools.Geometry}) described in \S\ref{app:tools}, retaining only the core perception tools (SAM3/DA3) and the scientific computing libraries (\texttt{numpy}, \texttt{scipy}) available in the execution environment. This configuration tests whether the agent can substitute pre-defined utility logic with on-the-fly numerical computation.
Variant \textbf{(I)} achieves performance on par with full \ours{}, suggesting that the persistent kernel with scientific primitives can largely compensate for the absent utility tools. 
We also report a no-perception variant \textbf{(II)}, which removes the perception tools (SAM3/DA3) from full \ours{}, leaving only the code-as-action interface with scientific libraries. The resulting $+2.7\%p$ gain over the no-tool baseline isolates the contribution of the action interface itself, independent of the perception tools.

\begin{wrapfigure}{r}{0.52\textwidth}
\centering
\includegraphics[width=\linewidth]{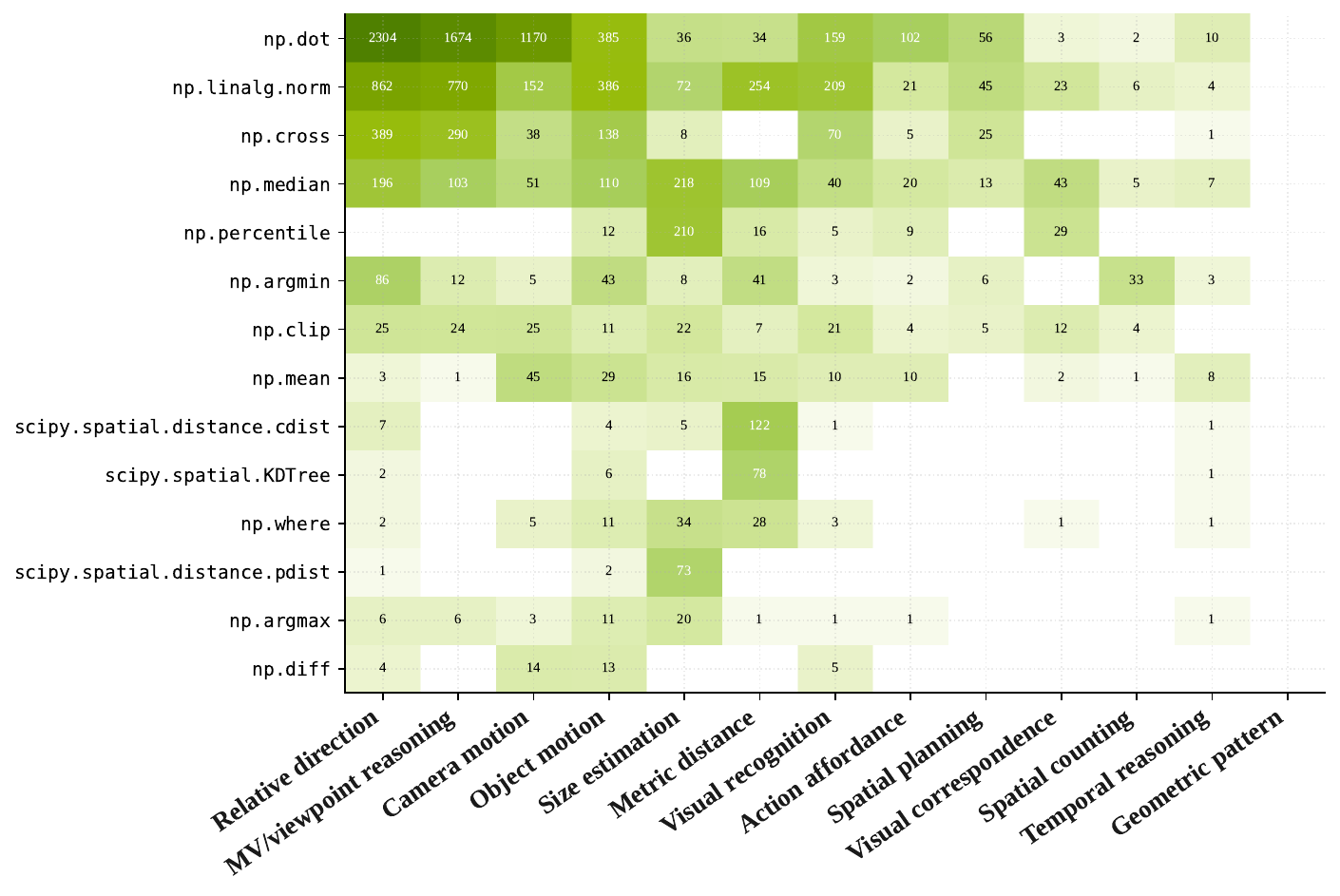}
\caption{\textbf{Composition adapts to the question type.} Primitive usage frequency across meta-categories.}
\label{fig:caa-vocab}
\end{wrapfigure}

\finding{The agent spontaneously adapts its tool composition to the question type: distance questions preferentially invoke KD-tree search and norm operations, while direction questions rely on dot products. (Fig.~\ref{fig:caa-vocab})}{}

To understand \emph{how} the agent composes tools, we analyze the distribution of primitives (i.e., numpy and scipy operations) invoked across meta-categories (Figure~\ref{fig:caa-vocab}). 
To enable a unified analysis across benchmarks, we group the fine-grained task categories of the 20 benchmarks into 13 semantically coherent meta-categories (e.g., depth estimation, relative direction, camera motion). 
The heatmap reveals a clear specialization: distance-type questions heavily use spatial indexing (\texttt{KDTree}) and vector norms, which are the natural building blocks for nearest-neighbor and proximity; direction-type questions instead rely on dot products and angular operations, which capture orientation and heading. Critically, this specialization is \emph{not} hard-coded; no category-specific prompt engineering or tool routing was applied. The agent selects geometrically appropriate primitives solely from the question semantics, demonstrating that \ours{}'s action interface unlocks a form of spontaneous, task-adaptive composition that is difficult to achieve with a structured tool-call interface.

\finding{\ours{} gains are largest precisely where chained geometric computation across frames and viewpoints is required, validating \ours{}'s action interface design. (Fig.~\ref{fig:pairwise_winloss_per_meta})}{}

In Figure~\ref{fig:pairwise_winloss_per_meta}, we compare \ours{} against the Structured tool-call and Single-pass Code baselines across 13 meta-categories. 
To construct these meta-categories, we map each benchmark's fine-grained category labels into one of 13 coarser groups spanning all 20 evaluated benchmarks.
\ours{} secures a net advantage in 11/13 categories over both Structured tool-call and Single-pass Code. 
The largest lifts ($+$6--9 pp) concentrate in Camera motion, Multi-view/viewpoint reasoning, and Relative direction, precisely the categories that require chained geometric computation across frames and viewpoints. This is where the persistent kernel, by enabling cross-step composition and revision, provides the greatest leverage. Where gains are smaller, the bottleneck is perception quality: Visual recognition tasks are already near-saturated by the backbone VLM, leaving little room for interface-level gains. Together, this breakdown confirms that the expressive action interface is the primary driver of performance, not model capacity or tool coverage.

\section{Conclusion}
We presented \ours{}, a training-free spatial reasoning agent that adopts code as the action interface, enabling a VLM to flexibly compose, inspect, and revise perception tool outputs across steps in a persistent Python kernel.
Evaluated across 20 spatial reasoning benchmarks and six VLM backbones from two model families, \ours{} achieves 59.9\% average accuracy and outperforms the recent spatial agent by \textbf{+11.2 points}, without any model- or benchmark-specific adaptation.
These results demonstrate that the design of the action interface is a highly impactful yet underexplored axis of improvement for spatial reasoning agents.

\section*{Acknowledgements}
We are deeply grateful to Valts Blukis, Siyi Chen, Yi Dong, Tsung-Yi Lin, Karan Sapra, Andrew Tao, Bowen Wen, and Zhiding Yu for their valuable discussions, insightful feedback, and generous support throughout this work.

\clearpage
\clearpage
\appendix

\section*{Supplementary Material}
\addcontentsline{toc}{section}{Supplementary Material}

\lstdefinestyle{promptStyle}{%
  basicstyle=\ttfamily\scriptsize,
  showstringspaces=false,
  breaklines=true,
  breakatwhitespace=false,
  breakindent=0pt,
  columns=fullflexible,
  upquote=true,
  aboveskip=4pt,
  belowskip=2pt,
  frame=single,
  framerule=0.4pt,
  framesep=4pt,
  backgroundcolor=\color{gray!5},
  linewidth=\linewidth,
  resetmargins=true,
}

\begingroup
\noindent
\vspace{0.6em}

\etocdepthtag.toc{appendix}
\etocsettagdepth{mainpaper}{none}
\etocsettagdepth{appendix}{subsection}
\etocsetnexttocdepth{subsection}
\etocsetstyle{section}
  {\setlength{\parindent}{0pt}\setlength{\parskip}{1pt}}
  {}
  {\noindent\makebox[1.6em][l]{\bfseries\etocnumber}\bfseries\etocname\dotfill\etocpage\par}
  {}
\etocsetstyle{subsection}
  {}
  {}
  {\noindent\hspace*{1.6em}{\normalfont\makebox[2.0em][l]{\etocnumber}\etocname}\dotfill\etocpage\par}
  {}
\etocsettocstyle{\noindent{\large\bfseries Contents}\par\vspace{0.3em}}{}
{\small\tableofcontents}
\endgroup
\clearpage

\section{Related Works}
\label{app:related}

\paragraph{Spatial reasoning in VLMs.}
Despite broad progress in vision-language models (VLMs), spatial reasoning remains a persistently limited capability~\citep{chen2024spatialvlm,cheng2024spatialrgpt,song2025robospatial}.
A common response is to fine-tune VLMs with spatial supervision, either by distilling 3D annotations into instruction data~\citep{chen2024spatialvlm,cheng2024spatialrgpt} or by augmenting the model with explicit geometry modules~\citep{hu2025g,fan2025vlm,zhang2026make}.
These approaches produce fast, self-contained models, but require retraining whenever perception modules or task distributions change.

\paragraph{Tool-augmented visual agents.}
Tool-augmented visual agents have an LLM compose calls to specialist vision modules~\citep{Gupta2022VisProg,surismenon2023vipergpt,shen2023hugginggpt,zhao2025pyvision,lu2026octotools}.
Early work synthesizes a full program in a single pass~\citep{Gupta2022VisProg,surismenon2023vipergpt}, subsequent work dispatches requests through structured tool menus~\citep{shen2023hugginggpt,lu2026octotools}.
These systems establish that external perception can extend an LLM beyond its native visual capabilities, but they either constrain the agent to a fixed tool interface that may limit generalization to novel compositions, or commit to a full program before any intermediate output can be inspected, or fuse perception and planning inside a single multimodal model that cannot independently verify intermediate tool outputs.

\paragraph{Spatial reasoning agents.}
GCA~\citep{chen2025geometrically} decouples the VLM into a semantic analyst that formalizes the query as geometric constraints and a task solver that executes tool calls within those bounds, RieMind~\citep{ropero2026riemind} grounds an LLM in an explicit 3D scene graph queried through typed geometric operations, and SpaceTools~\citep{chen2025spacetools} fine-tunes a VLM to coordinate a predefined set of perception tools through supervised demonstrations followed by interactive reinforcement learning.
Think3D~\citep{zhang2026think3d} uses a closely related but more specialized interface, repeatedly selecting camera viewpoints to render a reconstructed point cloud and reasoning over the rendered images.
pySpatial~\citep{luo2026pyspatial} composes reconstruction, camera-pose recovery, and novel-view synthesis into a 3D visual program, and VADAR~\citep{marsili2025visual} first synthesizes a task-specific Pythonic API and then a program that calls into it.
Neither writes code turn-by-turn in response to intermediate execution results, so per-step inspection of perception outputs is limited.

\paragraph{Code action interface for LLM agents.}
CodeAct~\citep{wang2024executable} showed that emitting executable Python code as the action outperforms JSON- and text-formatted action spaces for general-purpose LLM agents.
\ours{} instantiates this paradigm for spatial reasoning, contributing domain-specific design choices absent from general-purpose frameworks.
Where CodeAct focuses on the infrastructure of code execution, our contribution is the spatial-reasoning angle, namely controlled comparisons against alternative action interfaces and trace-level analyses that identify when code action interface helps and which spatial analysis patterns drive the gains (\S\ref{sec:abl_action}, \S\ref{sec:analysis}).

\section{Evaluation Protocol}
\label{app:benchmarks}

We evaluate \ours{} on the 20 spatial reasoning benchmarks reported in the main results, organized into five categories: single-image spatial reasoning (ERQA~\citep{team2025gemini}, Omni3D~\citep{marsili2025visual}, OmniSpatial~\citep{jia2025omnispatial}, SPBench~\citep{xu2025spbench}), multi-view spatial reasoning (MindCube~\citep{yin2025mindcube}, MMSI~\citep{yang2025mmsi}, SPAR-Bench~\citep{zhang2025from}), video spatial \& 4D reasoning (MMSI-Video~\citep{lin2025mmsivideo}, OSI-Bench~\citep{wu2025osibench}, PAI-Bench~\citep{zhou2025paibench}, VSI-Bench-U~\citep{brown2025shortcuts}, VSTI-Bench~\citep{fan2025vlm}, DSI-Bench~\citep{zhang2025dsibench}), general spatial reasoning (BLINK~\citep{fu2024blink}, SpatialTree~\citep{xiao2025spatialtree}, ViewSpatial~\citep{li2025viewspatial}), and general video understanding (CV-Bench~\citep{zhu2025cvbench}, PerceptComp~\citep{li2026perceptioncomp}, Video-MME~\citep{fu2025video}, Video-MME-v2~\citep{videommev2_2026}). 
Table~\ref{tab:benchmark_metrics} reports the per-sample scoring metric applied to each benchmark.

\begin{table}[h]
\caption{\textbf{Benchmark-specific evaluation metrics and scoring protocols.} Each benchmark mixes categorical questions (multiple-choice or binary) and numerical questions in different proportions. \texttt{Acc} denotes per-sample 1/0 categorical scoring, \texttt{MRA} denotes mean relative accuracy on numerical questions, and \texttt{VCI} denotes the SPAR-Bench metric used for the view change inference task (MRA averaged across five movement axes). The reported number for each benchmark is the unweighted mean of the per-sample scores.}
\label{tab:benchmark_metrics}
\centering
\setlength{\tabcolsep}{4pt}
\renewcommand{\arraystretch}{1.15}
{\small
\begin{tabular}{lll}
\toprule
\textbf{Benchmark} & \textbf{Category} & \textbf{Per-sample metric} \\
\midrule
ERQA~\citep{team2025gemini}             & Single-image spatial reasoning & Acc \\
Omni3D~\citep{marsili2025visual}        & Single-image spatial reasoning & Acc + MRA \\
OmniSpatial~\citep{jia2025omnispatial}  & Single-image spatial reasoning & Acc \\
SPBench~\citep{xu2025spbench}           & Single-image spatial reasoning & Acc + MRA \\
\midrule
MindCube~\citep{yin2025mindcube}        & Multi-view spatial reasoning & Acc \\
MMSI~\citep{yang2025mmsi}               & Multi-view spatial reasoning & Acc \\
SPAR-Bench~\citep{zhang2025from}        & Multi-view spatial reasoning & Acc + MRA + VCI \\
\midrule
MMSI-Video~\citep{lin2025mmsivideo}     & Video spatial \& 4D reasoning & Acc \\
OSI-Bench~\citep{wu2025osibench}        & Video spatial \& 4D reasoning & Acc + MRA \\
PAI-Bench~\citep{zhou2025paibench}      & Video spatial \& 4D reasoning & Acc \\
VSI-Bench-U~\citep{brown2025shortcuts}  & Video spatial \& 4D reasoning & Acc + MRA \\
VSTI-Bench~\citep{fan2025vlm}           & Video spatial \& 4D reasoning & Acc + MRA \\
DSI-Bench~\citep{zhang2025dsibench}     & Video spatial \& 4D reasoning & Acc \\
\midrule
BLINK~\citep{fu2024blink}               & General spatial reasoning & Acc \\
SpatialTree~\citep{xiao2025spatialtree} & General spatial reasoning & Acc + MRA \\
ViewSpatial~\citep{li2025viewspatial}   & General spatial reasoning & Acc \\
\midrule
CV-Bench~\citep{zhu2025cvbench}         & General video understanding & Acc \\
PerceptComp~\citep{li2026perceptioncomp}& General video understanding & Acc \\
Video-MME~\citep{fu2025video}           & General video understanding & Acc \\
Video-MME-v2~\citep{videommev2_2026}    & General video understanding & Acc \\
\bottomrule
\end{tabular}}
\end{table}

We apply a consistent per-sample scoring protocol across all benchmarks, using the same metric family while respecting each benchmark's original threshold settings. Categorical questions, including multiple-choice and binary questions, receive a score of $1$ when the predicted answer matches the reference and $0$ otherwise. Numerical questions are scored with mean relative accuracy (MRA)~\citep{yang2024think}, where the acceptance threshold follows each benchmark's originating implementation. SPAR-Bench's view-change-inference task uses VCI, defined as MRA averaged over its five movement axes. The reported number for each benchmark is the unweighted mean of the per-sample scores.

For benchmarks with more than $1,000$ samples, we evaluate a randomly chosen subset of $1,000$ samples drawn with a fixed seed; the same subset is reproduced by running the supplementary code with the same seed.

\clearpage
\section{Additional Analysis}
\begin{wrapfigure}{r}{0.5\textwidth}
\centering
\includegraphics[width=\linewidth]{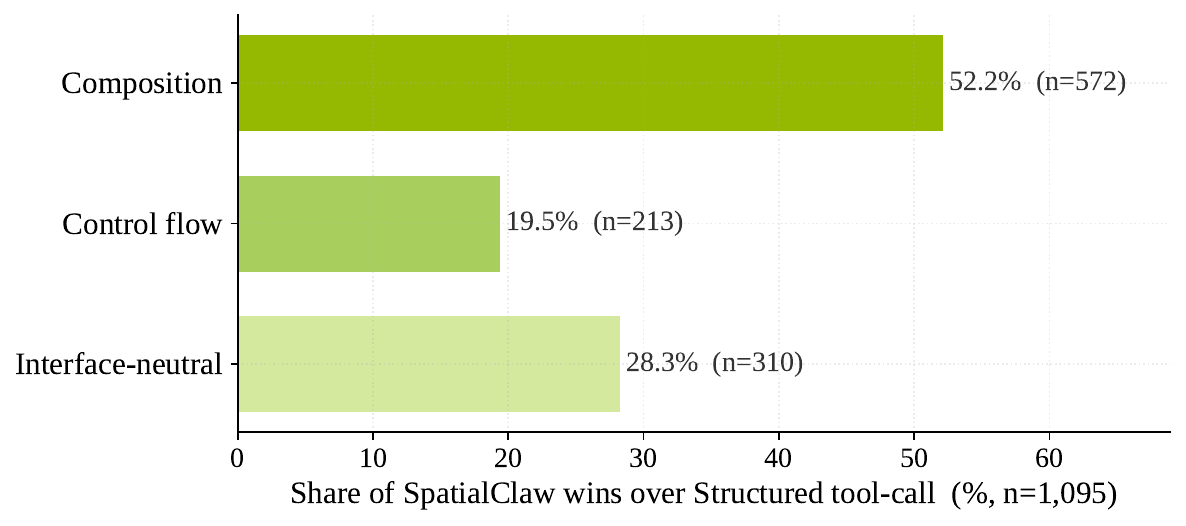}
\caption{\textbf{Attribution of \ours{}'s wins over structured tool-call via LLM-as-judge.} Over half of the gains are driven by \emph{code composition}, 19.5\% by \emph{control flow}, and 28.3\% are \emph{interface-neutral} wins on perceptual tasks unaffected by the action interface.}
\label{fig:caa-overview}
\end{wrapfigure}

\finding{Composition is the main driver of \ours{}'s gains over structured tool-call. (Fig.~\ref{fig:caa-overview})}{}

To identify the main driver of \ours{}'s gains over structured tool-call baseline, we examine the instances where \ours{} answers correctly but structured tool-call fails (Figure~\ref{fig:caa-overview}).
For each such instance, we provide an LLM judge (Gemini-3.1-Pro~\citep{team2023gemini}) with the full reasoning traces of both systems, the ground-truth answer, and the input images, and ask it to assign a binary label to each predefined attribution category; instances in which all categories receive a negative label are classified as \emph{interface-neutral}. 

We find that over 50\% of the instances are attributed to \emph{code composition} (i.e., chaining multiple tool calls into a single coherent program), while 19.5\% are attributed to \emph{control flow} (e.g., \texttt{if} or \texttt{for} statements that conditionally branch or iterate over intermediate results). 
The remaining 28.3\% are \emph{interface-neutral} wins, in which the correct answer depends on visual recognition or luck rather than the agent's action space, and either interface would have been equally capable. 
This is further causally supported by Tab.~\ref{tab:ablation}, which shows that performance degradation is minimal even when predefined utility functions are removed and replaced with on-the-fly numerical computation.

\finding{Geometric reasoning errors constitute a leading failure mode, with perception errors driven by VLM hallucinations and tool limitations as a notable secondary contributor.}{}

\begin{wrapfigure}{r}{0.5\textwidth}
\centering
\includegraphics[width=\linewidth]{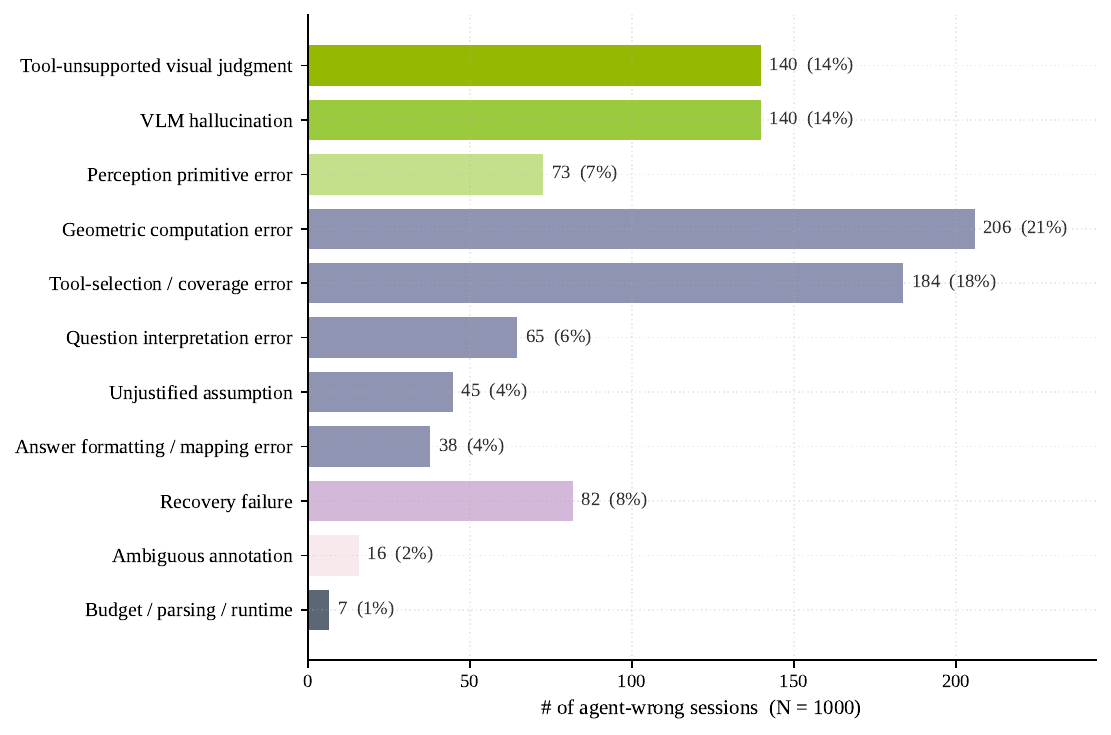}
\caption{\textbf{Failure-mode breakdown of incorrect agent sessions.} Each session is classified by an LLM-as-Judge (Gemini-3.1-Pro~\citep{team2023gemini}) into one of 11 fine-grained failure categories.}
\label{fig:caa-failure}
\end{wrapfigure}

In Fig.~\ref{fig:caa-failure}, we analyze the agent's failure modes by categorizing $1,000$ incorrect samples with LLM-as-Judge~\citep{zheng2023judging}, using a strong proprietary model (Gemini-3.1-Pro). For each sample, the judge is provided with the agent's full reasoning trace and the ground-truth answer, and assigns each session to one of 11 human-designed fine-grained categories.

On the perception side, the agent frequently attempts visual judgments that exceed the capabilities of its available tools, and its underlying VLM occasionally hallucinates objects, attributes, or spatial relations that are not actually present in the scene. Lower-level perception primitives such as detection and segmentation also produce localized errors that propagate into downstream reasoning. 
The predominant reasoning failures are geometric in nature, as the agent frequently errs in handling 3D coordinates, distances, angles, or projective relationships, even when the underlying perceptual evidence is sound. 
Closely related is a class of tool-selection and coverage failures, in which the agent invokes an inappropriate tool or omits a necessary intermediate step.
We additionally find cases of question misinterpretation, unjustified assumptions about unobserved scene properties, and inconsistencies between the internal computation and the final answer format. 
Beyond these direct errors, a non-trivial fraction of failures arises from the agent's inability to recover upon detecting an inconsistency, as it either commits to a flawed early hypothesis or oscillates without converging. 
The remaining residual cases involve ambiguous ground-truth annotations and occasional budget, parsing, or runtime issues, which together account for only a small share of observed failures. 

A promising direction for future work is to apply reinforcement learning to improve tool selection, coding of geometric operations, and error recovery within the agent loop.

\section{Baseline Implementations}
\label{app:baselines}

\S\ref{sec:interfaces} of the main paper compares the code as the action interface of \ours{} against two baselines, single-pass code and structured tool-calls. To isolate the effect of the interface, both baselines share the agent loop, the persistent kernel, the perception tools, and the planner; only the format of the per-step action differs, along with any adjustment to the per-sample step budget necessitated by the format.

\subsection{Single-Pass Code}
\label{app:baseline_single_pass}
The single-pass baseline collapses the multi-step loop to a single iteration. The agent receives the question, the key frames, and the same main-agent system prompt, and must produce one Python cell that calls perception tools, performs any required computation, and submits the answer through \texttt{ReturnAnswer}. Planning is disabled, and the workflow section of the system prompt is replaced by a single-turn variant that informs the agent there will be no further opportunity to act. The agent must therefore commit to a complete strategy before observing any intermediate mask, depth map, plot, or runtime error. When the cell does not reach \texttt{ReturnAnswer}, the same termination fallback used by the code as the action interface variant (\S\ref{app:error_handling}) returns a best-effort answer.

\subsection{Structured Tool-Calls}
\label{app:baseline_react}
The structured-tool-call baseline keeps the multi-step loop intact but constrains each step to a single named tool invocation expressed as JSON. The \emph{Code} section of the response envelope is replaced by a \emph{Tool Call} section that contains exactly one object of the form
\begin{lstlisting}[style=promptStyle]
{"tool": "tools.SAM3.segment_image_by_text",
 "args": {"image": "InputImages[0]", "prompt": "red car"}}
\end{lstlisting}
The specified tool call is first translated to Python code and then executed on the same kernel used by \ours{}, and its return value is stored in a named variable that later steps can reference. Arguments must be either literal values or references to previously bound names (i.e., the input variables, the tool catalogue, or the outputs of prior steps).
Arbitrary Python expressions are not allowed, and each step issues exactly one tool call, and intermediate results are accessible only through these named per-step variables.

\section{Agent System Design}
\label{app:system}

\subsection{System Configuration}
\label{app:topology}
\ours{} runs on two independently scalable serving roles. The first role hosts the language-model backbone and serves every LLM call the system makes, including the main agent, the planner, and the two isolated visual sessions invoked from inside the kernel (\apicode{vlm.locate} for coordinate grounding and \apicode{vlm.ask_with_thinking} for visual reasoning). The backbone is served via vLLM~\citep{kwon2023efficient}; multiple endpoints are reachable through a single OpenAI-compatible interface, and a session-aware router dispatches each call so that the prefix of the conversation lands on a consistent endpoint, exploiting the prefix cache that vLLM maintains across requests.

The second role hosts the perception models behind a lightweight HTTP service: Depth Anything 3~\citep{depthanything3} for 3D reconstruction and SAM3~\citep{carion2025sam} for segmentation. The kernel-side tools \apicode{tools.Reconstruct} and \apicode{tools.SAM3} are light-weight clients that call this service and convert the results into ordinary numpy arrays before returning to the agent.
For Depth Anything 3, we use the \texttt{DA3Nested-Giant-Large} variant, which pairs an any-view giant model with a metric depth model to reconstruct visual geometry at real-world metric scale.

This decomposition has two practical consequences. First, the language-model and perception backbones scale independently: the number of LLM nodes and perception nodes can each be tuned to their respective call demand without any changes to the agent code. Second, all LLM roles (agent, planner, grounding, and thinking) share the same backbone pool, so the main conversation and in-kernel visual queries share the same prefix cache and benefit from KV reuse.

\subsection{Input Preprocessing}
\label{app:input_preprocess}
All input images are resized such that the long edge does not exceed 768 pixels, preserving the original aspect ratio; this bounds the per-frame token cost. The agent VLM receives at most 32 frames as visual context: for video inputs longer than 32 frames, frames are drawn at uniform temporal intervals to yield 32 key frames; for multi-image inputs, the sequence is truncated to the first 32 images. The persistent kernel retains the full sampled frame sequence, allowing any frame to be revisited via \apicode{show()} or forwarded to a perception tool in subsequent steps. When a sample contains multiple videos, each is maintained as an independent frame sequence, enabling the agent to analyze and compare them independently.

\subsection{Persistent Kernel}
\label{app:kernel}
Each example is solved within a dedicated, stateful IPython kernel that persists throughout the entire inference run for that sample. Variables created during one cell (segmentation masks, reconstructions, depth and point arrays, intermediate visualizations, and partial numerical results) remain in scope for every subsequent cell, so the agent can revisit and recompose earlier evidence without re-issuing the underlying tool call. A per-cell wall-clock timeout protects the system from runaway code, and on repeated execution failures the kernel is restarted; when this happens, the input frames, sample metadata, the \texttt{tools} module, the \texttt{feedback} module, and \texttt{ReturnAnswer} are automatically re-injected so the agent resumes against an identical environment.

\subsection{Security Sandbox}
\label{app:sandbox}
Code emitted by the LLM is statically analyzed before execution. The analyzer traverses the Abstract Syntax Tree (AST) of each cell to identify and reject unsafe patterns, including file I/O, network access, dynamic-code primitives such as \texttt{exec} and \texttt{eval}, and direct imports of GPU-backend libraries. A complementary regular-expression pass catches patterns that may evade AST traversal, such as method-style writes (\texttt{.save}, \texttt{.to\_csv}). When a cell is rejected, the corresponding error message is delivered to the agent as feedback so it can revise the code within the same step, without adding latency to steps that pass inspection.

\subsection{Per-Frame Type Contract}
\label{app:per_frame}
Every frame-indexed perception output is encapsulated in a typed container that records the absolute frame indices to which it corresponds. Composing two such containers (e.g., a per-frame segmentation mask with a per-frame point map) triggers an automatic frame-index alignment check at the composition site; index mismatches raise an immediate exception identifying the offending frames. This mechanism guards against a class of silent semantic errors in which tensors are numerically well-formed but spatially misaligned, such as applying a mask derived from one frame to the point map of another. By surfacing the error at the point of composition, the violation appears directly in the agent's subsequent feedback rather than propagating silently to produce an incorrect final answer.

\subsection{Error Handling}
\label{app:error_handling}
Adopting code as the action interface exposes the system to a broader class of runtime failures than a fixed-API interface. \ours{} treats each such failure as an additional observation by routing the resulting exception or traceback directly into the agent's next feedback message, enabling the agent to diagnose and revise its code within the same episode rather than terminating on error.

\paragraph{Response-format errors.}
If the agent's response cannot be parsed into the four required fields (\emph{Purpose}, \emph{Reasoning}, \emph{Next Goal}, \emph{Code}), the validator records a \emph{format error} for the step. The malformed text is replaced in the conversation with a short placeholder that names the violation, and the agent is reminded to re-read the required format on the next turn. The malformed body is never echoed back, both to keep the context clean and to discourage the agent from imitating it.

\paragraph{Sandbox rejections.}
Code that the security analyzer rejects (\S\ref{app:sandbox}) is never executed. Instead, the rejection reason (identifying the forbidden module, builtin, or pattern) is returned as the step's error and surfaced to the agent in the next feedback message, exactly as if the code had failed at runtime. The agent therefore retries the same step with revised code rather than escalating.

\paragraph{Cell-execution exceptions.}
When a cell raises a Python exception, three steps are applied to trim the error output before the failure is forwarded to the agent.
The following example illustrates the condensed representation that is inserted into the conversation history after a faulting step:
\begin{lstlisting}[style=promptStyle]
**Purpose**: Perform 3D reconstruction and extract point clouds for the fireplace and the painting.
**Reasoning**: [errored -- condensed]
**Next Goal**: [errored -- condensed]
**Code**:
```python
# 1. 3D Reconstruction
recon = tools.Reconstruct.Reconstruct(InputImages)

# 2. Extract point clouds
# Fireplace is in InputImages[1]
fi_fire = seg_fireplace.frame_indices[0]  # <-- ERROR
# NameError: name 'seg_fireplace' is not defined
```
\end{lstlisting} 
First, the traceback is truncated to the final exception type, its message, and the offending source line; surrounding interpreter frames that reflect the execution infrastructure rather than the agent's code are discarded. Second, any code appearing after the faulting line is excluded from the context window, since it was never executed and therefore contributes no evidence to the agent's subsequent reasoning. Third, the \emph{Reasoning} and \emph{Next Goal} fields of the failed step are replaced by a condensed summary prior to insertion into the conversation history, preventing a single failure from disproportionately consuming the available context budget.

\paragraph{Cell timeouts.}
Each cell is subject to a per-cell wall-clock timeout to guard against non-terminating or excessively long-running code. Upon timeout, the kernel's user namespace is cleared and the execution environment is restored by re-injecting the input frames, sample metadata, and the \texttt{tools}, \texttt{feedback}, and \texttt{ReturnAnswer} entry points; the agent receives a timeout notification in its next observation and is expected to revise its code accordingly. If namespace restoration fails, the kernel process is fully restarted and the same re-injection sequence is applied.

\paragraph{Perception-tool retries.}
Calls to the GPU-backed perception tools may fail transiently due to server restarts or temporary overload. Each \apicode{tools.Reconstruct} or \apicode{tools.SAM3} call is automatically retried up to a fixed number of times, with increasing wait intervals between attempts; each retry also re-selects an available endpoint to route away from an unhealthy server. Only if all retries are exhausted is the error propagated to the agent as a Python exception. Similarly, transient network errors on the language-model side (e.g., connection drops or gateway timeouts) are distinguished from agent-side failures and do not count against the agent's error budget.

\paragraph{Hard budgets and forced termination.}
The loop carries explicit budgets on the number of LLM steps, the number of consecutive step-level failures, and the cumulative number of tool calls. When any budget is exceeded, the loop hands control to a termination node that is guaranteed to return an answer. The termination node attempts two fallback strategies in order: (i) a chain-of-thought fallback that prompts the VLM backbone to answer directly from the key frames using the same boxed-answer protocol as the no-tool baseline; and (ii) if the chain-of-thought fallback does not yield a parseable answer, a regex-based extraction over the agent's recent messages and variables. The agent therefore always returns a best-effort answer for any sample whose execution started, even when the persistent-kernel path was unable to reach \texttt{ReturnAnswer} on its own.

\subsection{Backbones}
\label{app:backbones}
\begin{table}[h]
\caption{\textbf{Backbones used in the main results.} All six are served through the same vLLM-based~\citep{kwon2023efficient} serving role and evaluated under identical configuration: the same system prompt, tool set, maximum step count, and input preprocessing across every benchmark.}
\label{tab:backbones}
\centering
\setlength{\tabcolsep}{6pt}
\renewcommand{\arraystretch}{1.15}
{\small
\begin{tabular}{lccc}
\toprule
\textbf{Backbone} & \textbf{Size} & \textbf{Quantization} & \textbf{Context} \\
\midrule
Qwen3.5-397B-A17B~\citep{qwen3.5}        & 397B (17B activated) & GPTQ Int4~\citep{frantar-gptq} & 262K \\
Qwen3.5-122B-A10B~\citep{qwen3.5}        & 122B (10B activated) & FP8       & 262K \\
Qwen3.6-35B-A3B~\citep{qwen36_35b_a3b}   & 35B  (3B activated)  & FP8       & 262K \\
Qwen3.6-27B~\citep{qwen3.6-27b}          & 27B                  & FP8       & 262K \\
Gemma4-31B~\citep{gemma4}                & 31B                  & FP8       & 256K \\
Gemma4-26B-A4B~\citep{gemma4}            & 26B  (4B activated)  & FP8       & 256K \\
\bottomrule
\end{tabular}}
\end{table}

We evaluate \ours{} with the six open-source backbones listed in Table~\ref{tab:backbones}. All six are served through the same vLLM-based serving role described in \S\ref{app:topology}. Crucially, every backbone is evaluated under \emph{identical} configuration: the same system prompt, the same set of perception tools, the same maximum step count, and the same input preprocessing. No prompt template, tool subset, or budget is tuned per benchmark or per backbone.

\section{Prompt Details}
\label{app:prompts}

\ours{} defines four VLM roles, each with its own system prompt: the main agent that writes code, the planner that produces a plan before any image is observed, and two isolated visual sessions invoked from inside the kernel through \apicode{vlm.locate} (coordinate grounding) and \apicode{vlm.ask_with_thinking} (visual reasoning). The number of calls per role varies per sample: the main agent is queried once per step, the planner is invoked once per sample, and the two kernel-invoked sessions are called as many times as the agent decides. This appendix summarizes what each prompt encodes.

\subsection{Main Agent Prompt}
\label{app:prompt_main}
The main agent's system prompt is organized into a fixed sequence of sections, which we describe in order. A short header introduces the role of the agent and names the input variables (\texttt{InputImages} and \texttt{Metadata}).

\paragraph{Response format.}
The first content section declares the response format: every reply must contain four markdown sections with the headings \emph{Purpose}, \emph{Reasoning}, \emph{Next Goal}, and \emph{Code}, where the code field holds a single Python cell to be executed in the kernel.

\paragraph{Visual access.}
The two following sections document the visual access surface available inside the kernel. The \apicode{show()} entry displays one or more images inline in the next feedback message, notes that \texttt{matplotlib} figures are auto-captured, and states the per-session image budgets. The \apicode{vlm.locate} entry calls an isolated grounding session that returns coordinates in a 0--1000 normalized scale and accepts up to eight images per call, and the \apicode{vlm.ask_with_thinking} entry calls an isolated reasoning session that returns a textual answer over up to 64 frames. The same section instructs the agent to treat \texttt{Not visible} and \texttt{Cannot determine from the images.}\ as valid responses and not to reassign the \texttt{vlm} or \texttt{feedback} variables.

\paragraph{Available tools.}
The \emph{Available Tools} section then documents the seven tool namespaces that the kernel exposes, with method signatures, argument types, expected input shapes, and short usage examples for each: \apicode{tools.Reconstruct}, \apicode{tools.SAM3}, \apicode{tools.Graph}, \apicode{tools.Time}, \apicode{tools.Mask}, \apicode{tools.Geometry}, and \apicode{tools.Draw}. The details can be found in \S\ref{app:tools}.

\paragraph{Coordinate systems.}
The coordinate-system section names four reference frames (pixel space, camera space, world space, and an object-centric frame defined by an object's facing direction), states the technical convention for camera-to-world matrices returned by \apicode{tools.Reconstruct} (4$\times$4 SE(3) matrices in OpenCV convention, with gravity-aligned world axes anchored to the first camera), and gives a code snippet that computes left/right/front/behind classifications by projecting the world-frame vector to a target onto the camera's forward and right axes.

\paragraph{Robust computation.}
The robust-computation section lists five principles: prefer median over mean for aggregations, compare across multiple frames before drawing conclusions, reason in metric units rather than pixels, sanity-check magnitudes against physical priors, and print numerical values before committing to a conclusion.

\paragraph{Cross-validation.}
The cross-validation section enumerates four complementary evidence sources (visual perception via \apicode{show()}, geometric computation, visualizations, and logical reasoning) and specifies a five-step diagnostic procedure that the agent follows when two of them disagree.

\paragraph{Kernel-side contract and budget.}
The remaining sections fix the kernel-side contract and the run-level budget. The \texttt{ReturnAnswer} section documents the single-call API used to submit a final answer and the accepted argument types (\texttt{str}, \texttt{int}, \texttt{float}). The code-rules section lists the modules pre-imported into the kernel, the modules and built-ins forbidden by the security sandbox, and the reserved names that the agent must not reassign (\texttt{feedback}, \texttt{tools}, \texttt{InputImages}, \texttt{Metadata}, \texttt{ReturnAnswer}, \texttt{show}, \texttt{RefImages}). The workflow section instructs the agent to follow the planner's plan and to confirm any spatial conclusion against at least two independent lines of evidence before calling \texttt{ReturnAnswer}, and is followed by a single line stating the per-sample step budget. A final session-input section describes the per-sample input variables, the frame-indexing convention for \texttt{InputImages}, the fields of \texttt{Metadata}, and the key-frame-to-variable mapping; an additional reference-images section is included when the sample contains inline \texttt{[reference image \#N]} tags.

\subsection{Planner Prompt}
\label{app:prompt_planner}
The planner is invoked once per sample, before the main agent begins execution. The planner receives the question text, the per-sample metadata (frame count, frame indices, and frame-rate fields), and the system prompt only.

\paragraph{Shared documentation.}
After a one-paragraph header that declares the planner's role and states that it does not see the actual frames, the prompt repeats the same \emph{Available Tools}, \apicode{show()}, \apicode{vlm.locate}, \apicode{vlm.ask_with_thinking}, and coordinate-system documentation that the main agent receives, and embeds the cross-validation and robust-computation sub-sections from the main agent prompt.

\paragraph{Planning strategy.}
The planning-strategy section maps question shapes to tool choices: questions about object coordinates point to \apicode{vlm.locate} followed by \apicode{tools.SAM3}; questions about three-dimensional geometry, depth, or camera pose point to \apicode{tools.Reconstruct}; questions answerable by visual judgment point to \apicode{show()}; and questions that require a textual reading over several frames, or a fallback when a specialized tool fails, point to \apicode{vlm.ask_with_thinking}. The same section gives explicit guidance on annotation overlays, on coordinate grounding through 0--1000 normalized pixel coordinates, and on the choice between quantitative computation and qualitative visual reading.

\paragraph{Task structure and rules.}
The session-input section reports the frame count and the frame-indexing convention and states that the planner cannot see the images, instructing it to plan an investigation rather than answer the question from the question text. The task section asks for six items in the output: a task analysis (including an explicit choice of coordinate system when the question is ambiguous), a list of information needs, an ordered computation plan, a verification checklist, a list of cross-validation steps, and a fallback plan. The critical-rules section states that the planner must respond in plain text only, must not include JSON outside the verification checklist, must not write executable Python code, and must not produce the answer or pre-conclude phrases such as ``the answer is likely \dots'' or ``I expect the answer is \dots''; the planning-strategy section adds that any hypothesis the planner forms from the question is itself what the tools must verify, rather than evidence in its own right.

\subsection{Vision Prompts}
\label{app:prompt_vision}
The two isolated visual sessions invoked from the kernel use their own system prompts.

\paragraph{Grounding session (\texttt{vlm.locate}).}
The grounding session called by \apicode{vlm.locate} specifies a three-step procedure: identify the exact object, annotation, or marker that the question describes; classify the image as PRESENT, ABSENT, or AMBIGUOUS with respect to that description; and answer accordingly. PRESENT cases produce coordinates in the format requested by the calling question. ABSENT and AMBIGUOUS cases reply with the literal string \texttt{Not visible} on its own line, optionally followed by one short clarifying sentence. The prompt includes three worked examples that illustrate each of the three classifications, and ends with a short formatting block stating that coordinates use the 0--1000 normalized scale, that the assistant follows a ``Reply with ONLY the numbers'' instruction when present, and that the same prompt also covers segmentation-overlay assessment and plot-and-chart reading.

\paragraph{Reasoning session (\texttt{vlm.ask\_with\_thinking}).}
The reasoning session called by \apicode{vlm.ask_with_thinking} specifies five rules: 1) ground every claim in what is directly observable in the supplied images, 2) use any frame indices referenced by the question in the answer, 3) reason carefully across the frames before producing a concise final answer, 4) reply with the literal string \texttt{Cannot determine from the images.}\ together with a one-line note when the supplied frames do not constrain the answer, and 5) treat each call as independent of any prior context.

\section{Tool API Reference}
\label{app:tools}

This appendix expands the one-paragraph tool description in the main paper into per-tool signatures and behaviors. The first two tools (\apicode{tools.Reconstruct} and \apicode{tools.SAM3}) are thin clients that call the perception serving role described in \S\ref{app:topology}; the remaining tools are pure-Python helper functions that execute in the same process as the kernel.

\subsection{\texttt{tools.Reconstruct}}
\label{app:tool_reconstruct}
\apicode{tools.Reconstruct} is a client wrapper for the perception serving role described in \S\ref{app:topology}. Its single entry method calls Depth Anything 3~\citep{depthanything3} on a frame batch and returns a \texttt{Reconstruction} object that exposes per-frame depth, camera intrinsics, camera-to-world extrinsics, a dense per-frame point map in world coordinates, and a top-down rendering helper function. The output coordinate system is gravity-aligned with respect to the first camera: $+Y$ points up, the ground plane sits at $Y \approx 0$, and the first camera looks toward $-Z$. The output spatial resolution matches the input frame resolution, so masks from \apicode{tools.SAM3} can be combined directly with the reconstructed point maps without resizing.

\paragraph{Method.}
\begin{itemize}[leftmargin=1.5em]
\item \apicode{Reconstruct(frames, frame_indices=None)}
  \begin{itemize}
  \item \emph{Input:} \apicode{frames} is a list of \apicode{FrameImage} or PIL (Python Imaging Library) images, with at most \apicode{reconstruct_max_frames} entries (set to 64 in our experiments). When \apicode{FrameImage} objects are passed, absolute video frame indices are auto-extracted from each frame's \apicode{frame_index} attribute and the \apicode{frame_indices} argument may be omitted.
  \item \emph{Output:} a \texttt{Reconstruction} object (described below).
  \end{itemize}
\end{itemize}

\paragraph{Reconstruction attributes.} The returned object is indexed by absolute frame index, not by position in the input list.
\begin{itemize}[leftmargin=1.5em]
\item \apicode{frame_indices}: \texttt{list[int]}, absolute frame indices in the same order as \texttt{frames}.
\item \texttt{num\_frames}: \texttt{int}; \texttt{metric\_scale}: \texttt{float}.
\item \texttt{depth[fi]}: \texttt{(H, W) float32} depth map at absolute frame \texttt{fi}.
\item \texttt{extrinsics[fi]}: \texttt{(4, 4) float64} camera-to-world SE(3) matrix in OpenCV convention.
\item \texttt{intrinsics[fi]}: dictionary with keys \texttt{fx}, \texttt{fy}, \texttt{cx}, \texttt{cy}.
\item \texttt{points[fi]}: \texttt{(H, W, 3) float32} dense point map in world coordinates.
\end{itemize}

\paragraph{render\_bev.}
\begin{itemize}[leftmargin=1.5em]
\item \apicode{recon.render_bev(masks=None, labels=None, ref_frame=0, ego_trajectory=True)}
  \begin{itemize}
  \item \emph{Input:} \texttt{masks} is a \texttt{PerFrameMask} or an \texttt{(N, $N_{\text{obj}}$, H, W)} boolean array; \texttt{labels} provides per-object names when a raw array is passed; \texttt{ref\_frame} is the absolute frame index of the reference camera; \texttt{ego\_trajectory} toggles drawing of the camera path.
  \item \emph{Output:} a \texttt{VisualFeedback} that the agent inspects with \apicode{show()}. Stationary objects are annotated with oriented bounding boxes and moving objects with colour-graded trajectory lines.
  \end{itemize}
\end{itemize}

\subsection{\texttt{tools.SAM3}}
\label{app:tool_sam3}
\apicode{tools.SAM3} is a client wrapper for the perception serving role and exposes single-frame and video-mode segmentation against SAM3~\citep{carion2025sam}, together with an object-existence check. All segmentation methods return a \texttt{PerFrameMask} whose frame indices are absolute video frame indices and align with the indices used by any \texttt{Reconstruction} the agent has built.

\paragraph{Image-mode methods.} Each operates on a single image and returns a \texttt{PerFrameMask} with one frame.
\begin{itemize}[leftmargin=1.5em]
\item \apicode{segment_image_by_text(image, prompt, label=None)}
  \begin{itemize}
  \item \emph{Input:} \texttt{image} is a PIL image; \texttt{prompt} is a free-form text description.
  \item \emph{Output:} \texttt{PerFrameMask} with one channel per detected instance.
  \end{itemize}
\item \apicode{segment_image_by_points(image, points, point_labels, label="object")}
  \begin{itemize}
  \item \emph{Input:} \texttt{points} is a list of \texttt{[x, y]} pixel coordinates and \texttt{point\_labels} is a parallel list with values $1$ (foreground) or $0$ (background).
  \item \emph{Output:} \texttt{PerFrameMask} for the prompted object.
  \end{itemize}
\item \apicode{segment_image_by_box(image, box, label="object")}
  \begin{itemize}
  \item \emph{Input:} \texttt{box} is \texttt{[x1, y1, x2, y2]} in pixels (xyxy convention).
  \item \emph{Output:} \texttt{PerFrameMask} for the boxed object.
  \end{itemize}
\end{itemize}

\paragraph{Video-mode methods.} Each propagates the segmentation across the requested frame range, with at most \texttt{sam3\_max\_video\_frames} frames per call. \texttt{start\_frame} and \texttt{end\_frame} are absolute video frame indices that select a temporal window; \texttt{prompt\_frame\_idx} is local to that window; \texttt{video\_index} is 1-indexed and selects the source video in multi-video samples.
\begin{itemize}[leftmargin=1.5em]
\item \apicode{segment_video_by_text(prompts, labels=None, prompt_frame_idx=0, start_frame=None, end_frame=None, video_index=1)}
  \begin{itemize}
  \item \emph{Input:} \texttt{prompts} is a list of text descriptions, one per object.
  \item \emph{Output:} \texttt{PerFrameMask} over the selected frame range.
  \end{itemize}
\item \apicode{segment_video_by_points(points_per_object, point_labels_per_object, labels, prompt_frame_idx=0, start_frame=None, end_frame=None, video_index=1)}
  \begin{itemize}
  \item \emph{Input:} per-object point clicks on the prompt frame, parallel foreground/background labels, and human-readable per-object labels.
  \item \emph{Output:} \texttt{PerFrameMask} over the selected frame range.
  \end{itemize}
\item \apicode{segment_video_by_box(boxes, labels, prompt_frame_idx=0, start_frame=None, end_frame=None, video_index=1)}
  \begin{itemize}
  \item \emph{Input:} per-object bounding boxes on the prompt frame and parallel human-readable labels.
  \item \emph{Output:} \texttt{PerFrameMask} over the selected frame range.
  \end{itemize}
\end{itemize}

\paragraph{Object-existence check.}
\begin{itemize}[leftmargin=1.5em]
\item \apicode{is_object_exist(images, object_name)}
  \begin{itemize}
  \item \emph{Input:} \texttt{images} is a list of PIL images and \texttt{object\_name} is a text description.
  \item \emph{Output:} a dictionary with keys \texttt{exists} (list of booleans), \texttt{counts} (list of instance counts per image), and \texttt{summary} (a short text summary).
  \end{itemize}
\end{itemize}

\paragraph{PerFrameMask object.} The returned object exposes \apicode{frame_indices}, \texttt{labels}, \texttt{num\_frames}, and \texttt{num\_objects}, and supports the following access patterns at absolute frame index \texttt{fi}.
\begin{itemize}[leftmargin=1.5em]
\item \texttt{seg.get\_mask(frame=fi, object=k)}: 2D boolean mask of object $k$ (\texttt{k} may be an index or a string label).
\item \texttt{seg[fi]}: \texttt{($N_{\text{obj}}$, H, W)} boolean stack at frame \texttt{fi}.
\item \texttt{seg.get\_centroid\_3d(recon, frame=fi, object=k)}: median 3D world position of the masked points, or \texttt{None}.
\item \texttt{seg.get\_masked\_points(recon, frame=fi, object=k)}: \texttt{(K, 3)} array of masked world points.
\item \texttt{seg.visualize(fi)}: \texttt{VisualFeedback} overlay for verification with \apicode{show()}.
\end{itemize}

\subsection{\texttt{tools.Geometry}}
\label{app:tool_geometry}
\apicode{tools.Geometry} is a static class that provides the numerical primitives most commonly needed for spatial reasoning over reconstructed scenes. All methods operate on \texttt{numpy} arrays.

\paragraph{Methods.}
\begin{itemize}[leftmargin=1.5em]
\item \apicode{euclidean_distance(p1, p2)}
  \begin{itemize}
  \item \emph{Input:} two 1D arrays of length~3.
  \item \emph{Output:} \texttt{float}, the Euclidean distance.
  \end{itemize}
\item \apicode{angle_between_vectors(v1, v2)}
  \begin{itemize}
  \item \emph{Input:} two 3D vectors.
  \item \emph{Output:} \texttt{float}, the angle in degrees.
  \end{itemize}
\item \apicode{project_point_to_camera(point_3d, c2w, fx, fy, cx, cy)}
  \begin{itemize}
  \item \emph{Input:} a 3D world point, a $4 \times 4$ camera-to-world SE(3) matrix, and the four scalar intrinsics.
  \item \emph{Output:} \texttt{(u, v)} pixel coordinates, or \texttt{None} when the point is behind the camera.
  \end{itemize}
\item \apicode{rotation_matrix_from_vectors(v_from, v_to)}
  \begin{itemize}
  \item \emph{Input:} two 3D vectors (need not be unit length).
  \item \emph{Output:} a $3 \times 3$ rotation matrix that aligns \texttt{v\_from} with \texttt{v\_to}.
  \end{itemize}
\item \apicode{transform_points(points, matrix)}
  \begin{itemize}
  \item \emph{Input:} a \texttt{(..., 3)} array of points and a $4 \times 4$ SE(3) matrix.
  \item \emph{Output:} a transformed array with the same shape as the input.
  \end{itemize}
\item \apicode{fit_ground_plane_ransac(points, confidence, conf_threshold=0.3, n_iterations=1000, inlier_threshold=0.05)}
  \begin{itemize}
  \item \emph{Input:} a \texttt{(H, W, 3)} or \texttt{(N, 3)} point cloud and a matching confidence array; thresholds and iteration count are optional.
  \item \emph{Output:} \texttt{(plane\_normal, inlier\_mask)} via RANSAC~\citep{fischler1981random}, or \texttt{(None, None)} when fitting fails. The sign of \texttt{plane\_normal} is unspecified and is disambiguated by the caller.
  \end{itemize}
\item \apicode{normalized_to_pixel(coords, width, height)}
  \begin{itemize}
  \item \emph{Input:} a sequence of coordinates in the 0--1000 normalized scale used by \apicode{vlm.locate}, together with the image width and height in pixels.
  \item \emph{Output:} a \texttt{list[float]} of pixel coordinates.
  \end{itemize}
\end{itemize}

\subsection{\texttt{tools.Mask}}
\label{app:tool_mask}
\apicode{tools.Mask} is a static class that provides standard mask statistics over the boolean mask arrays returned by \apicode{tools.SAM3}. All inputs are 2D \texttt{numpy} arrays with shape \texttt{(H, W)} unless otherwise noted.

\paragraph{Methods.}
\begin{itemize}[leftmargin=1.5em]
\item \texttt{centroid(mask)}
  \begin{itemize}
  \item \emph{Input:} a 2D boolean mask.
  \item \emph{Output:} \texttt{(cx, cy)} pixel coordinates of the median; \texttt{(nan, nan)} when the mask is empty.
  \end{itemize}
\item \texttt{centroids(masks)}
  \begin{itemize}
  \item \emph{Input:} an \texttt{(N, H, W)} batch of boolean masks.
  \item \emph{Output:} a list of \texttt{(cx, cy)} centroids, one per element of the batch.
  \end{itemize}
\item \texttt{area(mask)}
  \begin{itemize}
  \item \emph{Input:} a 2D boolean mask.
  \item \emph{Output:} \texttt{int}, the number of \texttt{True} pixels.
  \end{itemize}
\item \apicode{bounding_box(mask)}
  \begin{itemize}
  \item \emph{Input:} a 2D boolean mask.
  \item \emph{Output:} \texttt{(x1, y1, x2, y2)} in pixels, or \texttt{None} when the mask is empty. For masks with more than 100 \texttt{True} pixels, the box is computed from the 1st and 99th percentiles to discard outliers.
  \end{itemize}
\item \texttt{iou(mask\_a, mask\_b)}
  \begin{itemize}
  \item \emph{Input:} two boolean masks of identical shape.
  \item \emph{Output:} \texttt{float}, the intersection over union.
  \end{itemize}
\item \texttt{intersection(mask\_a, mask\_b)}
  \begin{itemize}
  \item \emph{Input:} two boolean masks of identical shape.
  \item \emph{Output:} a boolean mask of the same shape, the element-wise AND.
  \end{itemize}
\item \apicode{mask_to_bbox(mask)}
  \begin{itemize}
  \item \emph{Input:} a 2D boolean mask.
  \item \emph{Output:} a 4-element \texttt{numpy} array \texttt{[x1, y1, x2, y2]}, or \texttt{None} when the mask is empty.
  \end{itemize}
\end{itemize}

\subsection{\texttt{tools.Time}}
\label{app:tool_time}
\apicode{tools.Time} is a utility that converts between frame indices and seconds. It is initialized from the input metadata's frame rate and total frame count, and is meaningful only when \texttt{Metadata.is\_video} is true and \texttt{Metadata.fps} is set.

\paragraph{Methods.}
\begin{itemize}[leftmargin=1.5em]
\item \apicode{frame_to_seconds(frame_index)}
  \begin{itemize}
  \item \emph{Input:} an integer or float frame index.
  \item \emph{Output:} \texttt{float}, the corresponding time in seconds.
  \end{itemize}
\item \apicode{seconds_to_frame(seconds)}
  \begin{itemize}
  \item \emph{Input:} a number of seconds.
  \item \emph{Output:} \texttt{int}, the nearest frame index, clamped to the valid range.
  \end{itemize}
\item \apicode{frame_range_to_seconds(start_frame, end_frame)}
  \begin{itemize}
  \item \emph{Input:} two frame indices.
  \item \emph{Output:} \texttt{float}, the duration in seconds between them.
  \end{itemize}
\item \apicode{get_frame_at_time(seconds)}: alias for \texttt{seconds\_to\_frame}.
\end{itemize}

\subsection{\texttt{tools.Graph}}
\label{app:tool_graph}
\apicode{tools.Graph} is a utility that wraps \texttt{matplotlib} to produce diagnostic line plots from numerical sequences. The plot is returned as a \texttt{VisualFeedback} that the agent inspects inline with \apicode{show()} and that also carries a short text summary (minimum, maximum, mean, and trend) accessible as \texttt{chart.description}.

\paragraph{Method.}
\begin{itemize}[leftmargin=1.5em]
\item \apicode{plot(values, validity=None, x_label="Frame", y_label="Value", title=None)}
  \begin{itemize}
  \item \emph{Input:} \texttt{values} is a 1D \texttt{numpy} array of length~$T$. \texttt{validity} is an optional 1D boolean array of the same length whose \texttt{False} entries are rendered as gaps. \texttt{x\_label}, \texttt{y\_label}, and \texttt{title} are display strings.
  \item \emph{Output:} a \texttt{VisualFeedback} containing the rendered plot and a text description.
  \end{itemize}
\end{itemize}

\subsection{\texttt{tools.Draw}}
\label{app:tool_draw}
\apicode{tools.Draw} is a static class that produces PIL-based annotation overlays. All methods accept a \texttt{numpy} \texttt{(H, W, 3) uint8} array, a PIL image, or an \texttt{InputImages[i]} entry, and return a PIL image of the same size; the input image is never modified. Colours may be passed as an \texttt{(R, G, B)} \texttt{uint8} tuple or as any \texttt{matplotlib} or CSS4 colour name. Each method accepts either a single primitive or a list of primitives, and a matching list of colours when per-primitive colouring is desired.

\paragraph{Methods.}
\begin{itemize}[leftmargin=1.5em]
\item \apicode{draw_bbox(image, bboxes, colors=None, thickness=None)}
  \begin{itemize}
  \item \emph{Input:} \texttt{bboxes} is a single \texttt{(x1, y1, x2, y2)} in pixels or a list of such boxes.
  \item \emph{Output:} a PIL image with axis-aligned rectangle outlines drawn.
  \end{itemize}
\item \apicode{draw_line(image, lines, colors=None, thickness=None)}
  \begin{itemize}
  \item \emph{Input:} \texttt{lines} is a single \texttt{(x1, y1, x2, y2)} in pixels or a list of such segments.
  \item \emph{Output:} a PIL image with line segments drawn.
  \end{itemize}
\item \apicode{draw_point(image, points, colors=None, radius=None)}
  \begin{itemize}
  \item \emph{Input:} \texttt{points} is a single \texttt{(x, y)} in pixels or a list of such points.
  \item \emph{Output:} a PIL image with filled circles drawn.
  \end{itemize}
\end{itemize}

\section{Limitations and Broader Impact}
\label{app:limitations}

\subsection{Limitations}
\label{app:limitations_sub}
The main remaining bottleneck of \ours{} is the perceptual quality of the backbone vision-language model and of the perception tools it composes. The failure-mode analysis in the main paper attributes the largest share of remaining errors to perception rather than to the action interface, which means that further interface design has diminishing returns at the current scale of evaluation. We expect the principal axis for future improvement to be perceptual quality, both in the backbone vision-language model and in the underlying reconstruction and segmentation models.

\subsection{Broader Impact}
\label{app:broader_impact}
\ours{} is training-free and adds no parameters to the backbone vision-language model. Existing models can therefore be extended with stronger spatial reasoning ability without additional training data and without fine-tuning, which is particularly valuable in domains where data collection is expensive or impractical, including robotics, embodied applications, and assistive systems. Because the same configuration transfers across backbones and benchmarks without modification, the framework increases the practical value of vision-language models that are already deployed.

\clearpage
\setcitestyle{numbers}
\bibliographystyle{plainnat}
\bibliography{main}

\end{document}